\documentclass{article}
\usepackage{graphicx}
\usepackage{subfigure}
\usepackage{amsfonts,amssymb,amsmath}
\usepackage{color}
\usepackage{url}
\usepackage{tablefootnote}
\usepackage{mathrsfs}
\usepackage{algorithm}
\usepackage{algorithmic}
\usepackage{natbib}
\usepackage[paper=a4paper,textwidth=170mm,textheight=260mm]{geometry}

\newcommand*\samethanks[1][\value{footnote}]{\footnotemark[#1]}

\newtheorem{proposition}{Proposition}

\newcommand{\argmin}{\operatornamewithlimits{argmin}}

\title{Dictionary learning for fast classification based on soft-thresholding}

\author{Alhussein Fawzi\thanks{Ecole Polytechnique Federale de Lausanne (EPFL), Signal Processing Laboratory (LTS4), Lausanne 1015-Switzerland. Email: (alhussein.fawzi@epfl.ch, pascal.frossard@epfl.ch)} \and
Mike Davies\thanks{IDCOM, The University of Edinburgh, Edinburgh, UK. Email: mike.davies@ed.ac.uk}
\and Pascal Frossard\samethanks[1]}

\begin{document}

\maketitle

\begin{abstract}

Classifiers based on sparse representations have recently been shown to provide excellent results in many visual recognition and classification tasks. However, the high cost of computing sparse representations at test time is a major obstacle that limits the applicability of these methods in large-scale problems, or in scenarios where computational power is restricted. We consider in this paper a simple yet efficient alternative to sparse coding for feature extraction. We study a classification scheme that applies the \textit{soft-thresholding} nonlinear mapping in a dictionary, followed by a linear classifier. 
A novel supervised dictionary learning algorithm tailored for this low complexity classification architecture is proposed.
The dictionary learning problem, which \textit{jointly} learns the dictionary and linear classifier, is cast as a \textit{difference of convex} (DC) program and solved efficiently with an iterative DC solver. We conduct experiments on several datasets, and show that our learning algorithm that leverages the structure of the classification problem outperforms generic learning procedures. Our simple classifier based on soft-thresholding also competes with the recent sparse coding classifiers, when the dictionary is learned appropriately. 
The adopted classification scheme further requires less computational time at the testing stage, compared to other classifiers. The proposed scheme shows the potential of the adequately trained soft-thresholding mapping for classification and paves the way towards the development of very efficient classification methods for vision problems.

\end{abstract}

\section{Introduction}
\label{sec:intro}


The recent decade has witnessed the emergence of huge volumes of high dimensional information produced by all sorts of sensors. For instance, a massive amount of high-resolution images are uploaded on the Internet every minute. In this context, one of the key challenges is to develop techniques to process these large amounts of data in a computationally efficient way. We focus in this paper on the \textit{image classification} problem, which is one of the most challenging tasks in image analysis and computer vision. Given training examples from multiple classes, the goal is to find a rule that permits to predict the class of test samples. \textit{Linear classification} is a computationally efficient way to categorize test samples. It consists in finding a linear separator between two classes. 

Linear classification has been the focus of much research in statistics and machine learning for decades and the resulting algorithms are well understood. However, many datasets cannot be separated linearly and require complex nonlinear classifiers. 
A popular nonlinear scheme, which leverages the efficency and simplicity of linear classifiers, embeds the data into a high dimensional feature space, where a linear classifier is eventually sought. The feature space mapping is chosen to be nonlinear in order to convert nonlinear relations to linear relations.
This nonlinear classification framework is at the heart of the popular kernel-based methods \citep{shawe2004kernel} that make use of a computational shortcut to bypass the explicit computation of feature vectors. Despite the popularity of kernel-based classification, its computational complexity at test time strongly depends on the number of training samples \citep{burges1998tutorial}, which limits its applicability in large scale settings. 

A more recent approach for nonlinear classification is based on \textit{sparse coding}, which consists in finding a compact representation  of the data in an overcomplete dictionary. Sparse coding is known to be beneficial in signal processing tasks such as denoising \citep{elad2006image}, inpainting \citep{fadili2009inpainting}, coding \citep{figueras2006}, but it has also recently emerged in the context of classification, where it is viewed as a nonlinear feature extraction mapping. 
It is usually followed by a linear classifier \citep{raina2007self}, but can also be used in conjunction with other classifiers \citep{wright2009robust}. 
Classification architectures based on sparse coding have been shown to work very well in practice and even achieve state-of-the-art results on particular tasks \citep{mairal2012task, yang2009linear}.
The crucial drawback of sparse coding classifiers is however the prohibitive cost of computing the sparse representation of a signal or image sample at test time. This limits the relevance of such techniques in large-scale vision problems or when computational power is scarce.

\begin{figure}[t]
\centering
\includegraphics[width=0.5\textwidth]{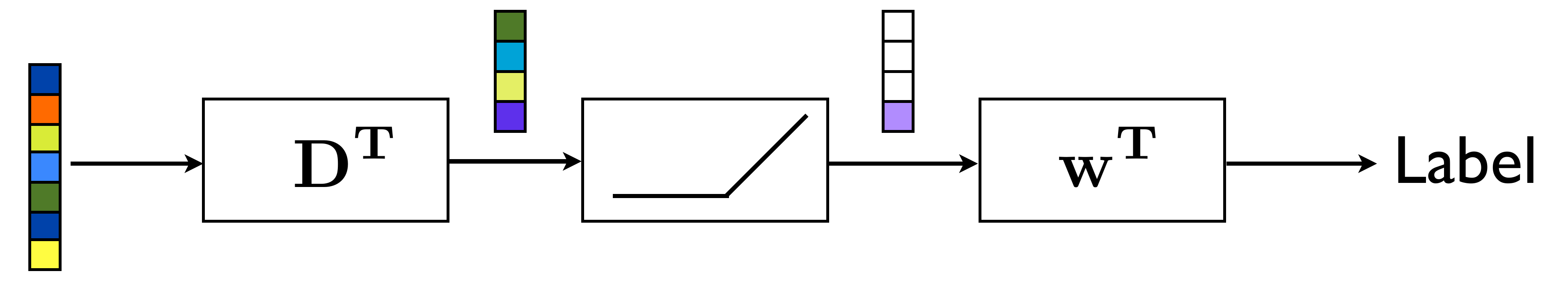}
\caption{\label{fig:last_classifier}  Soft-thresholding classification scheme. The box in the middle applies the soft-thresholding non-linearity $h_{\alpha}$.}
\end{figure}

To remedy to these large computational requirements, we adopt in the classification a computationally efficient sparsifying transform, the soft thresholding mapping $h_{\alpha}$, defined by:
\begin{align}
\label{eq:soft_threshold}
h_{\alpha} (z) = \max(0, z-\alpha)  \triangleq (z - \alpha)_+,
\end{align}
for $\alpha \in \mathbb{R}_+$ and $(\cdot)_+ = \max(0, \cdot)$. Note that, unlike the usual definition of soft-thresholding given by $\text{sgn}(z) (|z| - \alpha)_+$, we consider here the \textit{one-sided} version of the soft-thresholding map, where the function is equal to zero for negative values (see Fig. \ref{fig:soft_thresholding} (a) vs. Fig \ref{fig:soft_thresholding} (b)). The map $h_{\alpha}$ is naturally extended to vectors $\mathbf{z}$ by applying the scalar map to each coordinate independently.
Given a dictionary $\mathbf{D}$, this map can be applied to a transformed signal $\mathbf{z} = \mathbf{D^T x}$ that represents the coefficients of features in a signal $\mathbf{x}$. Its outcome, which only considers the most important features of $\mathbf{x}$, is used for classification. In more details, we consider in this paper the following simple two-step procedure for classification:
\begin{enumerate}
\item \textbf{Feature extraction: } Let $\mathbf{D} = [\mathbf{d_1} | \dots | \mathbf{d_N}]\in \mathbb{R}^{n \times N}$ and $\alpha \in \mathbb{R}_+$. Given a test point $\mathbf{x} \in \mathbb{R}^n$, compute $h_{\alpha} (\mathbf{D^T x})$.
\item \textbf{Linear classification: }  Let $\mathbf{w} \in \mathbb{R}^N$. If $\mathbf{w}^T h_{\alpha} (\mathbf{D^T x})$ is positive, assign $\mathbf{x}$ to class $1$. Otherwise, assign to class $-1$.
\end{enumerate}
The architecture is illustrated in Fig. \ref{fig:last_classifier}. 
The proposed classification scheme has the advantage of being simple, efficient and easy to implement as it involves a single matrix-vector multiplication and a $\max$ operation. 
The soft-thresholding map has been successfully used in \citep{coates2011importance}, as well as in a number of deep learning architectures \citep{kavukcuoglu2010learning}, which shows the relevance of this efficient feature extraction mapping. 
The remarkable results in \citet{coates2011importance} show that this simple encoder, when coupled with a standard learning algorithm, can often achieve results comparable to those of sparse coding, provided that the number of labeled samples and the dictionary size are large enough. However, when this is not the case, a proper training of the classifier parameters $(\mathbf{D}, \mathbf{w})$ becomes crucial for reaching good classification performance. This is the objective of this paper.


We propose a novel \textit{supervised dictionary learning} algorithm, which we call LAST (Learning Algorithm for Soft-Thresholding classifier). It \textit{jointly} learns the dictionary  $\mathbf{D}$ and the linear classifier $\mathbf{w}$ tailored for the classification architecture based on soft-thresholding. 
We pose the learning problem as an optimization problem comprising a loss term that controls the classification accuracy and a regularizer that prevents overfitting. This problem is shown to be a  \textit{difference-of-convex} (DC)  program, which is solved efficiently with an iterative DC solver.
We then perform extensive experiments on textures, digits and natural images datasets, and show that the proposed classifier, coupled with our dictionary learning approach, exhibits remarkable performance with respect to numerous competitor methods. In particular, we show that our classifier provides comparable or better classification accuracy than sparse coding schemes. 

The rest of this paper is organized as follows. In the next Section, we highlight the related work. In Section \ref{sec:problem_formulation}, we formulate the dictionary learning problem for classifiers based on soft-thresholding. Section \ref{sec:dc_optimization} then presents our novel learning algorithm, LAST, based on DC optimization. In Section \ref{sec:exp_results}, we perform extensive experiments on textures, natural images and digits datasets and Section \ref{sec:discussion} finally gathers a number of important observations on the dictionary learning algorithm, and the classification scheme.

\section{Related work}
\label{sec:related_work}



We first highlight in this section the difference between the proposed approach and existing techniques from the sparse coding and dictionary learning literature. Then, we draw a connection between the considered approach and neural network models on the architecture and optimization aspects. 

%
%

\subsection{Sparse coding}

The classification scheme adopted in this paper shares similarities with the now popular architectures that use sparse coding at the feature extraction stage. We recall that the sparse coding mapping, applied to a datapoint $\mathbf{x}$ in a dictionary $\mathbf{D}$ consists in solving the optimization problem
\begin{align}
\argmin_{\mathbf{c} \in \mathbb{R}^N} \| \mathbf{x} - \mathbf{D c} \|_2^2 + \lambda \| \mathbf{c} \|_1.
\label{eq:sparse_coding}
\end{align}
It is now known that, when the parameters of the sparse coding classifier are trained in a discriminative way, excellent classification results are obtained in many vision tasks \citep{mairal2012task, mairal2008supervised, ramirez2010classification}. In particular, significant gains over the standard reconstructive dictionary learning approaches are obtained when the dictionary is optimized for classification. Several dictionary learning methods also consider an additional structure (e.g., low-rankness) on the dictionary, in order to incorporate a task-specific prior knowledge \citep{zhang2013learning, chen2012low, ma2012sparse}. This line of research is especially popular in face recognition applications, where a mixture of subspace model is known to hold \citep{wright2009robust}. Up to our knowledge, all the discriminative dictionary learning methods optimize the dictionary in regards to the sparse coding map in Eq. (\ref{eq:sparse_coding}), or a variant that still requires to solve a non trivial optimization problem. In our work however, we introduce a discriminative dictionary learning method \textit{specific to the efficient soft-thresholding map}. Interestingly, soft-thresholding can be viewed as a coarse approximation to non-negative sparse coding, as we show in Appendix \ref{sec:softthresh_sparsecoding}. This further motivates the use of soft-thresholding for feature extraction, as the merits of sparse coding for classification are now well-established.

Closer to our work, several approaches have been introduced to approximate sparse coding with a more efficient feed-forward predictor \citep{kavukcuoglu2010fast, gregor2010learning}, whose parameters are learned in order to minimize the approximation error with respect to sparse codes. These works are however different from ours in several aspects. First, our approach does not require the result of the soft-thresholding mapping to be close to that of sparse coding. We rather require solely a good classification accuracy on the training samples. Moreover, our dictionary learning approach is purely supervised, unlike \citet{kavukcuoglu2010fast, kavukcuoglu2010learning}. Finally, these methods often use nonlinear maps (e.g., hyperbolic tangent in \citet{kavukcuoglu2010fast}, multi-layer soft-thresholding in \citet{gregor2010learning}) that are different from the one considered in this paper. 
The single soft-thresholding mapping considered here has the advantage of being simple, very efficient and easy to implement in practice. It is also strongly tied to  sparse coding (see Appendix \ref{sec:softthresh_sparsecoding}).

\subsection{Neural networks}
\label{sec:connection_nn}

\begin{figure}[ht]
\centering
\includegraphics[width=0.5\textwidth]{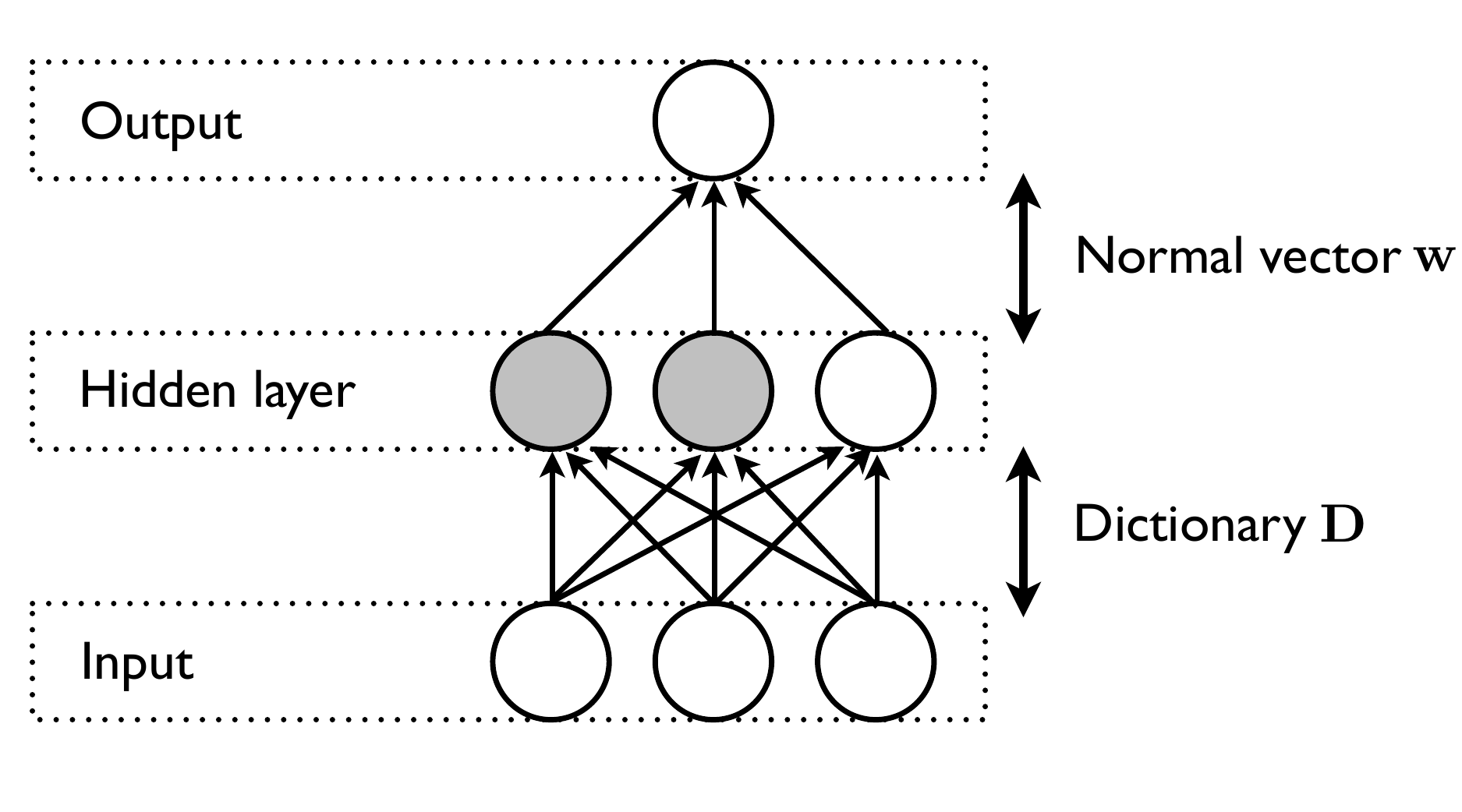}
\caption{\label{fig:nn_fig}  Neural network representation of our classification architecture. Greyed neurons have zero activation value.}
\end{figure}

The classification architecture considered in our work is also quite strongly related to artificial neural network models \citep{Bishop1995}. Neural network models are multi-layer architectures, where each layer consists of a set of neurons. The neurons compute a linear combination of the activation values of the preceding layer, and an \textit{activation function} is then used to convert the neurons' weighted input to its activation value. Popular choices of activation functions are logistic sigmoid and hyperbolic tangent nonlinearities. Our classification architecture can be seen as a neural network with one hidden layer and $h_{\alpha}$ as the hidden units' activation function, and zero bias (Fig. \ref{fig:nn_fig}). Equivalently, the activation function can be set to $\max(0, x)$ with a constant bias $-\alpha$ across all hidden units. The dictionary $\mathbf{D}$ defines the connections between the input and hidden layer, while $\mathbf{w}$ represents the weights that connect the hidden layer to the output.

In an important recent contribution, \citet{glorot2011deep} showed that using the rectifier activation function $\max(0, x)$ results in better performance for deep networks than the more classical hyperbolic tangent function. On top of that, the rectifier nonlinearity is more biologically plausible, and leads to sparse networks; a property that is highly desirable in representation learning \citep{bengio2013representation}. While the architecture considered in this paper is close to that of \citet{glorot2011deep}, it differs in several important aspects. 
First, our architecture assumes that hidden units have a bias equal to $-\alpha < 0$, shared across all the hidden units, while it is unclear whether any constraint on the bias is set in the existing rectifier networks. The parameter $\alpha$ is intimately related to the sparsity of the features. This can be justified by the fact that $h_{\alpha}$ is an approximant to the non-negative sparse coding map with sparsity penalty $\alpha$ (see Appendix \ref{sec:softthresh_sparsecoding}). Without imposing any restriction on the neurons' bias (e.g., negativity) in rectifier networks, the representation might however not be sparse. This potentially explains the necessity to use an additional $\ell_1$ sparsifying regularizer on the activation values in \citet{glorot2011deep} to enforce the sparsity of the network, while sparsity is achieved implicitly in our scheme.
Second, unlike the work of \citep{glorot2011deep} that employs a biological argument to introduce the rectifier function, we choose the soft-thresholding nonlinearity due to its strong relation to sparse coding. Our work therefore provides an independent motivation for considering the rectifier activation function, while the biological motivation in \citep{glorot2011deep} in turn gives us another motivation for considering soft-thresholding. Third, rectified linear units are very often used in the context of deep networks \citep{maas2013rectifier,zeiler2013rectified}, and seldom used with only one hidden layer. In that sense, the classification scheme considered in this paper has a simpler description, and can be seen as a particular instance of the general neural network models. 

From an optimization perspective, our learning algorithm leverages the simplicity of our classification architecture and is very different from the generic techniques used to train neural networks. In particular, while neural networks are generally trained with stochastic gradient descent, we adopt an optimization based on the DC framework that directly exploits the structure of the learning problem. 

\section{Problem formulation}
\label{sec:problem_formulation}
We present below the learning problem, that estimates jointly the dictionary $\mathbf{D} \in \mathbb{R}^{n \times N}$ and linear classifier $\mathbf{w} \in \mathbb{R}^N$ in our fast classification scheme described in Section \ref{sec:intro}.
We consider the binary classification task where $\mathbf{X} = [\mathbf{x_1} | \dots | \mathbf{x_m}] \in \mathbb{R}^{n \times m}$ and $\mathbf{y} = [y_1 | \dots | y_m ] \in \{ -1, 1 \}^m$ denote respectively the set of training points and their associated labels. We consider the following supervised learning formulation
\begin{align}
\label{eq:prob_formulation}
\argmin_{\mathbf{D}, \mathbf{w}} \sum_{i=1}^m L (y_i \mathbf{w^T} h_{\alpha} (\mathbf{D^T x_i})) + \frac{\nu}{2} \| \mathbf{w} \|_2^2,
\end{align}
where 
$L$ denotes a convex loss function that penalizes incorrect classification of a training sample and $\nu$ is a regularization parameter that prevents overfitting. The soft-thresholding map $h_\alpha$ has been defined in Eq. (\ref{eq:soft_threshold}). Typical loss functions that can be used in Eq. (\ref{eq:prob_formulation}) are the hinge loss ($L(x) = \max(0, 1 - x)$), which we adopt in this paper, or its smooth approximation, the logistic loss ($L(x) = \log(1+e^{-x})$). The above optimization problem attempts to find a dictionary $\mathbf{D}$ and a linear separator $\mathbf{w}$ such that $\mathbf{w^T} (\mathbf{D^T x_i} - \alpha)_+$ has the same sign as $y_i$ on the training set, which leads to correct classification. At the same time, it keeps $\| \mathbf{w} \|_2$ small in order to prevent overfitting. Note that to simplify the exposition, the bias term in the linear classifier is dropped. However, our study extends straightforwardly to include nonzero bias.

The problem formulation in Eq. (\ref{eq:prob_formulation}) is reminiscent of the popular support vector machine (SVM) training procedure, where only a linear classifier $\mathbf{w}$ is learned. Instead, we embed the nonlinearity directly in the problem formulation, and learn jointly the dictionary $\mathbf{D}$ and the linear classifier $\mathbf{w}$. This significantly broadens the applicability of the learned classifier to important nonlinear classification tasks. Note however that adding a nonlinear mapping raises an important optimization challenge, as the learning problem is no more convex.

When we look closer at the optimization problem in Eq. (\ref{eq:prob_formulation}), we note that, for any $\alpha > 0$, the objective function is equal to:
\begin{align*}
& \sum_{i=1}^m L ( y_i \alpha \mathbf{w^T} h_1 ( \mathbf{D^T x_i} / \alpha )  ) + \frac{\nu}{2} \| \mathbf{w} \|_2^2 \\
= & \sum_{i=1}^m L ( y_i \mathbf{\tilde{w}^T} h_1 ( \mathbf{\tilde{D}^T x_i} )  ) + \frac{\nu'}{2} \| \mathbf{\tilde{w}} \|_2^2,
\end{align*}
where $\tilde{\mathbf{w}} = \alpha \mathbf{w}$, $\tilde{\mathbf{D}} = \mathbf{D} / \alpha$ and $\nu' = \nu / \alpha^2$. Therefore, without loss of generality, we set the sparsity parameter $\alpha$ to $1$ in the rest of this paper. This is in contrast with traditional dictionary learning approaches based on $\ell_0$ or $\ell_1$ minimization problems, where a sparsity parameter needs to be set manually beforehand. Fixing $\alpha = 1$ and unconstraining the norms of the dictionary atoms essentially permits to adapt the sparsity to the problem at hand. 
This represents an important advantage, as setting the sparsity parameter is in general a difficult task. A sample $\mathbf{x}$ is then assigned to class `$+1$' if $\mathbf{w^T} h_{1} (\mathbf{D^T x})  > 0$, and class `$-1$' otherwise.

Finally, we note that, even if our focus primarily goes to the binary classification problem, the extension to multi-class can be easily done through a one-vs-all strategy, for instance.

\section{Learning algorithm}
\label{sec:dc_optimization}

The problem in Eq. (\ref{eq:prob_formulation}) is non-convex and difficult to solve in general. In this section, we propose to relax the original optimization problem and cast it as a \textit{difference-of-convex} (DC) program. Leveraging this property, we introduce LAST, an efficient algorithm for learning the dictionary and the classifier parameters in our classification scheme based on soft-thresholding.



\subsection{Relaxed formulation}
\label{sec:relaxed_formulation}
We rewrite now the learning problem in an appropriate form for optimization. We start with a simple but crucial change of variables. Specifically, we define $\mathbf{u_j} \leftarrow |w_j| \mathbf{d_j}$, $v_j \leftarrow |w_j|$ and $s_j \leftarrow \text{sgn} (w_j)$. Using this change of variables, we have for any $1 \leq i \leq m$, 
\begin{align*}
y_i \mathbf{w^T} h_1(\mathbf{D^T x_i}) & = y_i \sum_{j=1}^N \text{sgn} (w_j) (|w_j|\mathbf{d_j^T x_i} - |w_j|)_+ \\
																   & = y_i \sum_{j=1}^N s_j (\mathbf{u_j^T x_i} - v_j)_+.
\end{align*}
Therefore, the problem in Eq.(\ref{eq:prob_formulation}), with $\alpha = 1$, can be rewritten in the following way:
\begin{align}
\label{eq:prob_formulation_Uvs}
\argmin_{\mathbf{U}, \mathbf{v}, \mathbf{s}} & \sum_{i=1}^m L \left( y_i \sum_{j=1}^N s_j (\mathbf{u_j^T x_i} - v_j)_+ \right) + \frac{\nu}{2} \| \mathbf{v} \|_2^2, \\
& \text{ subject to } \mathbf{v} > 0 \nonumber.
\end{align}

The equivalence between the two problem formulations in Eqs. (\ref{eq:prob_formulation}) and (\ref{eq:prob_formulation_Uvs}) only holds when 
the components of the linear classifier $\mathbf{w}$ are restricted to be all non zero. This is however not a limiting assumption as zero components in the normal vector of the optimal hyperplane of Eq. (\ref{eq:prob_formulation}) can be removed, which is equivalent to using a dictionary of smaller size.

The variable $\mathbf{s}$, that is the sign of the components of $\mathbf{w}$, essentially encodes the ``classes'' of the different atoms. In other words, an atom $\mathbf{d_j}$ for which $s_j = +1$ (i.e., $w_j$ is positive) is most likely to be active for samples of class `$1$'. Conversely, atoms with $s_j = -1$ are most likely active for class `$-1$' samples. We assume here that the vector $\mathbf{s}$ is known a priori. In other words, this means that we have a prior knowledge on the proportion of class $1$ and class $-1$ atoms in the desired dictionary. For example, setting half of the entries of the vector $\mathbf{s}$ to be equal to $+1$ and the other half to $-1$ encodes the prior knowledge that we are searching for a dictionary with a balanced number of class-specific atoms. Note that $\mathbf{s}$ can be estimated from the distribution of the different classes in the training set, assuming that the proportion of class-specific atoms in the dictionary should approximately follow that of the training samples.


\begin{figure}[ht]
\centering
\subfigure[]{
\includegraphics[width=0.22\textwidth]{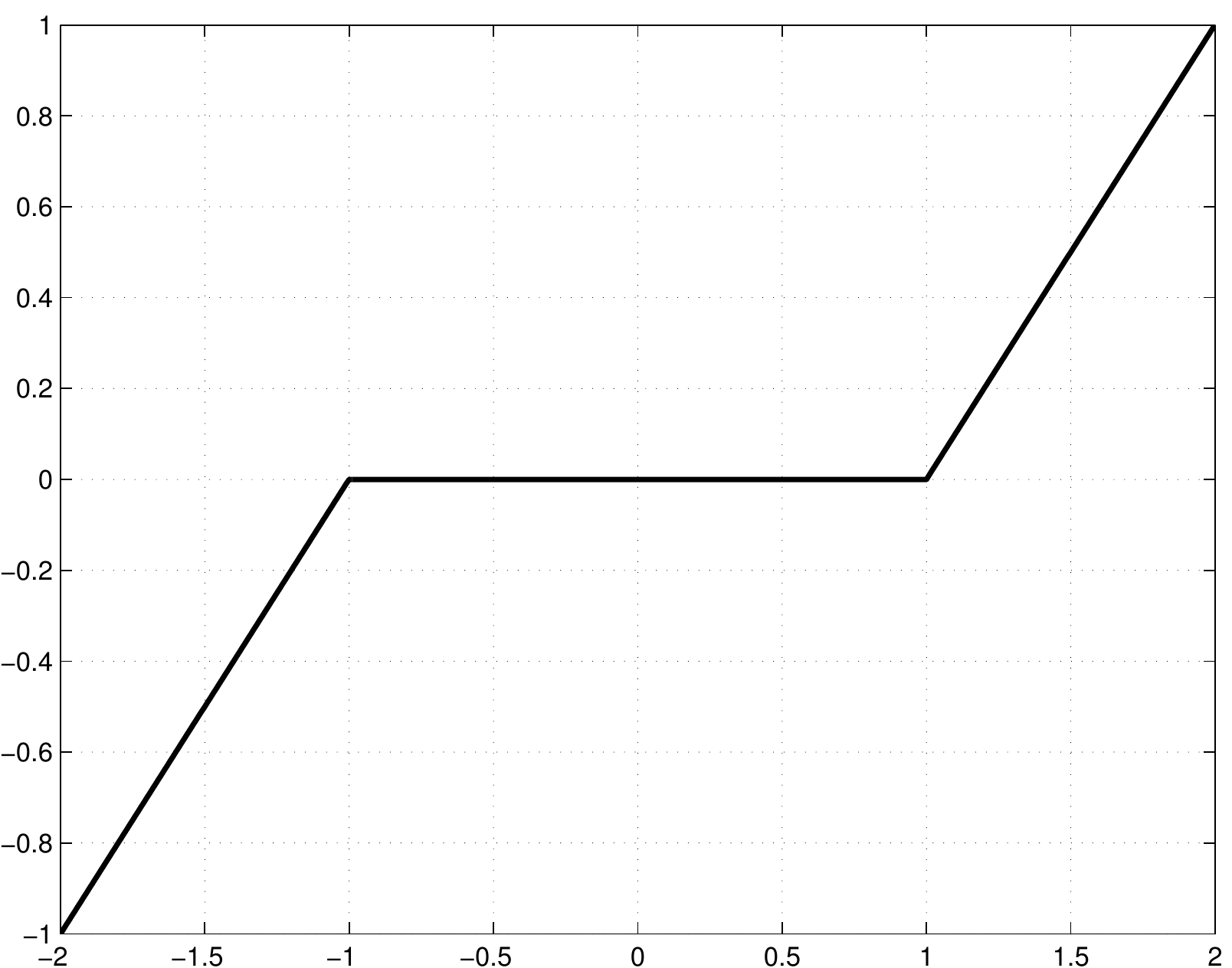}
}
\subfigure[]{
\includegraphics[width=0.22\textwidth]{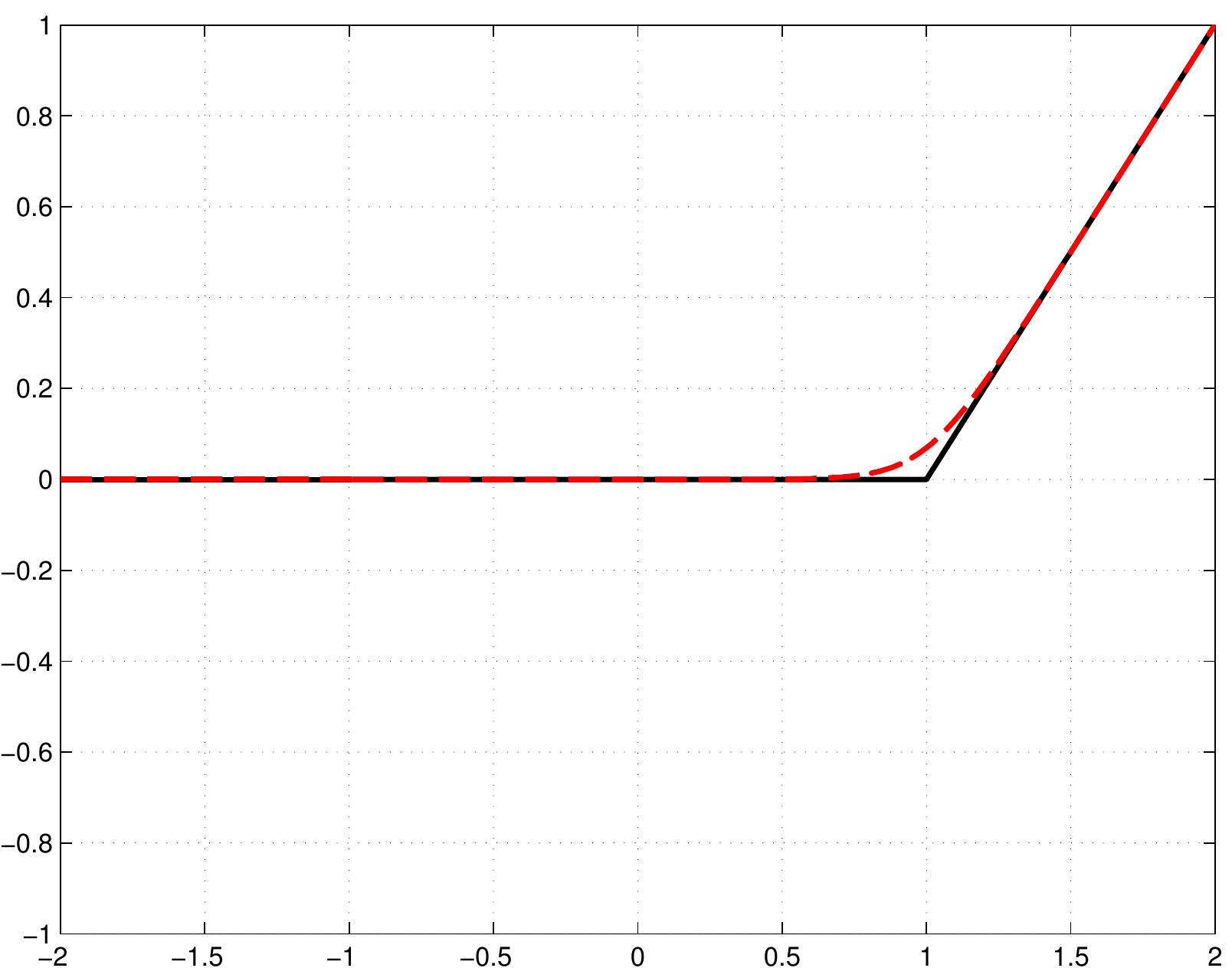}
}
\caption{\label{fig:soft_thresholding} (a): $\text{sgn}(x) (|x| - \alpha)_+$, (b): $h_{\alpha}$  (solid), and its smooth approximation $q(x-\alpha)$ (dashed), with $\beta = 10$. We used $\alpha = 1$.}
\end{figure}

After the above change of variables, we now approximate the term $(\mathbf{u_j^T x_i} - v_j)_+$ in Eq.(\ref{eq:prob_formulation_Uvs}) with a smooth function $q(\mathbf{u_j^T x_i} - v_j)$ where $q(x) = \frac{1}{\beta} \log \left( 1 + \exp \left( \beta x \right) \right)$, and $\beta$ is a parameter that controls the accuracy of the approximation (Fig. \ref{fig:soft_thresholding} (b)). Specifically, as $\beta$ increases, the quality of the approximation becomes better. The function $q$ with $\beta = 1$ is often referred to as ``soft-plus'' and plays an important role in the training objective of many classification schemes, such as the classification restricted Boltzmann machines \citep{larochelle2012learning}.
Note that this approximation is used only to make the optimization easier at the learning stage; at test time, the original soft-thresholding is applied for feature extraction.

Finally, we replace the strict inequality $\mathbf{v} > 0$ in Eq. (\ref{eq:prob_formulation_Uvs}) with $\mathbf{v} \geq \epsilon$, where $\epsilon$ is a small positive constant number. The latter constraint is easier to handle in the optimization, yet both constraints are essentially equivalent in practice.

We end up with the following optimization problem:
\begin{align}
\text{(P)}: 
\begin{array}{ll}
\displaystyle \argmin_{\mathbf{U}, \mathbf{v}} & \displaystyle \sum_{i=1}^m L \left( y_i \sum_{j=1}^N s_j q(\mathbf{u_j^T x_i} - v_j) \right) + \frac{\nu}{2} \| \mathbf{v} \|_2^2, \\
& \text{ subject to } \mathbf{v} \geq \epsilon \nonumber,
\end{array}
\end{align}
that is a relaxed version of the learning problem in Eq. (\ref{eq:prob_formulation_Uvs}).
Once the optimal variables $(\mathbf{U}, \mathbf{v})$ are determined, $\mathbf{D}$ and $\mathbf{w}$ can be obtained using the above change of variables. 





\subsection{DC decomposition}

The problem (P) is still a nonconvex optimization problem that can be hard to solve using traditional methods, such as gradient descent or Newton-type methods.  However, we show in this section that problem (P) can be written as a \textit{difference of convex} (DC) program \citep{horst2000introduction} which leads to efficient solutions. 

We first define DC functions. A real-valued function $f$ defined on a convex set $U \subseteq \mathbb{R}^n$ is called DC on $U$ if, for all $\mathbf{x} \in U$, $f$ can be expressed in the form 
\begin{align*}
f(\mathbf{x}) = g(\mathbf{x}) - h(\mathbf{x}),
\end{align*}
where $g$ and $h$ are convex functions on $U$. A representation of the above form is said to be a DC decomposition of $f$. Note that DC decompositions are clearly not unique, as $f(\mathbf{x}) = (g(\mathbf{x}) + c(\mathbf{x})) - (h(\mathbf{x}) + c(\mathbf{x}))$ provides other decompositions of $f$, for any convex function $c$. Optimization problems of the form $\min_\mathbf{x} \{ f(\mathbf{x}): f_i(\mathbf{x}) \leq 0, i = 1, \dots, p \}$, where $f$ and $f_i$ for $1 \leq i \leq p$ are all DC functions, are called \textit{DC programs}.

The following proposition now states that the problem (P) is DC:
\begin{proposition}
\label{prop:dc_nature}
For any convex loss function $L$ and any convex function $q$, the problem $(P)$ is DC.
\end{proposition}

While Proposition \ref{prop:dc_nature} states that the problem (P) is DC, it does not give an explicit decomposition of the objective function, which is crucial for optimization. The following proposition exhibits a decomposition when $L$ is the hinge loss.

\begin{proposition}
\label{prop:dc_decomposition}
When $L(x) = \max(0, 1 - x)$, the objective function of problem (P) is equal to $g-h$, where
\[
\begin{array}{lll}
g & = \frac{\nu}{2} \| \mathbf{v} \|_2^2 + \displaystyle \sum_{i=1}^m \max \Bigl( & \displaystyle \sum_{j: s_j = y_i} q(\mathbf{u_j^T x_i} - v_j), \\
& & 1 + \displaystyle \sum_{j: s_j \neq y_i} q(\mathbf{u_j^T x_i} - v_j) \Bigr), \\
h & \multicolumn{2}{l}{= \displaystyle \sum_{i=1}^m \displaystyle \sum_{j: s_j = y_i} q(\mathbf{u_j^T x_i} - v_j).}
\end{array}
\]
\end{proposition}

The proofs of Propositions \ref{prop:dc_nature} and \ref{prop:dc_decomposition} are given in Appendix \ref{sec:proof_dc_decomp}. Due to Proposition \ref{prop:dc_decomposition}, the problem (P) can be solved efficiently using a DC solver.

\subsection{Optimization}

DC problems are well studied optimization problems and efficient optimization algorithms have been proposed in \citep{horst2000introduction, tao1998dc} with good performance in practice (see \citet{tao2005dc} and references therein, \citet{sriperumbudur2007sparse}). While there exists a number of popular approaches that solve \textit{globally} DC programs (e.g., cutting plane and branch-and-bound algorithms \citep{horst2000introduction}), these techniques are often inefficient and limited to very small scale problems. A robust and efficient difference of convex algorithm (DCA) is proposed in \citet{tao1998dc}, which is suited for solving general large scale DC programs. DCA is an iterative algorithm that consists in solving, at each iteration, the convex optimization problem obtained by linearizing $h$ (i.e., the non convex part of $f = g - h$) around the current solution. The local convergence of DCA is proven in Theorem 3.7 of \citet{tao1998dc}, and we refer to this paper for further theoretical guarantees on the stability and robustness of the algorithm. Although DCA is only guaranteed to reach a local minima, the authors of \citet{tao1998dc} state that DCA often converges to a global optimum. When this is not the case, using multiple restarts might be used to improve the solution. We note that DCA is very close to the concave-convex procedure (CCCP) introduced in \citep{yuille2002concave}.

At iteration $k$ of DCA, the linearized optimization problem is given by:
\begin{align}
\argmin_{(\mathbf{U}, \mathbf{v})} \{ g(\mathbf{U}, \mathbf{v}) - Tr(\mathbf{U}^T \mathbf{A}) - \mathbf{v}^T \mathbf{b} \} \text{ subject to } \mathbf{v} \geq \epsilon.
\label{eq:each_iter_dca}
\end{align}
where $\mathbf{(A, b)} = \nabla h(\mathbf{U^k}, \mathbf{v^k})$ and $(\mathbf{U^k}, \mathbf{v^k})$ are the solution estimates at iteration $k$, and the functions $g$ and $h$ are defined in Proposition \ref{prop:dc_decomposition}. Note that, due to the convexity of $g$, the problem in Eq. (\ref{eq:each_iter_dca}) is convex and can be solved using any convex optimization algorithm \citep{boyd2004convex}. The method we propose to use here is a projected first-order stochastic subgradient descent algorithm. Stochastic gradient descent is an efficient optimization algorithm that can handle large training sets \citep{akata2013good}. To make the exposition clearer, we first define the function:
\[
\begin{array}{lll}
p(\mathbf{U}, \mathbf{v}; \mathbf{x_i}, y_i) & = \max \Bigl( & \displaystyle \sum_{j: s_j = y_i} q(\mathbf{u_j^T x_i} - v_j), \\
   & & 1 + \displaystyle \sum_{j: s_j \neq y_i} q(\mathbf{u_j^T x_i} - v_j) \Bigr) \\
   & \multicolumn{2}{l}{+ \frac{1}{m} \left( \frac{\nu}{2} \| \mathbf{v}\|_2^2 - Tr(\mathbf{U}^T \mathbf{A}) - \mathbf{v}^T \mathbf{b} \right) }.
\end{array}
\]
The objective function of Eq. (\ref{eq:each_iter_dca}) that we wish to minimize can then be written as $\sum_{i=1}^m p(\mathbf{U}, \mathbf{v}; \mathbf{x_i}, y_i)$. We solve this optimization problem with the projected stochastic subgradient descent algorithm in Algorithm \ref{alg:algo_sgd}.

\begin{algorithm}[ht]
\begin{algorithmic}
\STATE $\mathbf{1.}$ Initialization: $\mathbf{U} \leftarrow \mathbf{U^k}$ and $\mathbf{v} \leftarrow \mathbf{v^k}$.
\STATE $\mathbf{2.}$ For $t = 1, \dots, T$
\STATE \hspace{2mm}$\mathbf{2.1}$ Let $(\mathbf{x}, y)$ be a randomly chosen training point, and its associated label.
\STATE \hspace{2mm}$\mathbf{2.2}$ Choose the stepsize $\rho_t \leftarrow \min(\rho, \rho \frac{t_0}{t})$.
\STATE \hspace{2mm}$\mathbf{2.3}$ Update $\mathbf{U}$, and $\mathbf{v}$, by projected subgradient step:
\begin{align*}
\mathbf{U} & \leftarrow \mathbf{U} - \rho_t \displaystyle \partial_{\mathbf{U}} p(\mathbf{U}, \mathbf{v}; \mathbf{x}, y), \\
\mathbf{v} & \leftarrow \displaystyle \Pi_{\mathbf{v} \geq \epsilon} \left( \mathbf{v} - \rho_t \displaystyle \partial_{\mathbf{v}} p(\mathbf{U}, \mathbf{v}; \mathbf{x}, y) \right),
\end{align*}
where $\Pi_{\mathbf{v} \geq \epsilon}$ is the projection operator on the set $\mathbf{v} \geq \epsilon$.
\STATE $\mathbf{3.}$ Return $\mathbf{U^{k+1}} \leftarrow \mathbf{U}$ and $\mathbf{v^{k+1}} \leftarrow \mathbf{v}$.
\end{algorithmic}
\caption{\label{alg:algo_sgd} Optimization algorithm to solve the linearized problem in Eq. (\ref{eq:each_iter_dca})}
\end{algorithm}

In more details, at each iteration of Algorithm \ref{alg:algo_sgd}, a training sample $(\mathbf{x}, y)$ is drawn. $\mathbf{U}$ and $\mathbf{v}$ are then updated by performing a step in the direction $\partial p(\mathbf{U}, \mathbf{v}; \mathbf{x}, y)$. Many different stepsize rules can be used with stochastic gradient descent methods. In this paper, similarly to the strategy employed in \citet{mairal2012task}, we have chosen a stepsize that remains constant for the first $t_0$ iterations, and then takes the value $\rho t_0 / t$.\footnote{The precise choice of the parameters $\rho$ and $t_0$ are discussed later in Section \ref{sec:parameter_selection}.} Moreover, to accelerate the convergence of the stochastic gradient descent algorithm, we consider a small variation of Algorithm \ref{alg:algo_sgd}, where a minibatch containing several training samples along with their labels is drawn at each iteration, instead of a single sample. This is a classical heuristic in stochastic gradient descent algorithms. 
Note that, when the size of the minibatch is equal to the number of training samples, this algorithm reduces to traditional batch gradient descent.


Finally, our complete LAST learning algorithm based on DCA is formally given in Algorithm \ref{alg:algo_DCA}. Starting from a feasible point $\mathbf{U^0}$ and $\mathbf{v^0}$, LAST solves iteratively the constrained convex problem given in Eq. (\ref{eq:each_iter_dca}) with the solution proposed in Algorithm \ref{alg:algo_sgd}. Recall that this problem corresponds to the original DC program (P), except that the function $h$ has been replaced by its linear approximation around the current solution $(\mathbf{U^k}, \mathbf{v^k})$ at iteration $k$. 
Many criteria can be used to terminate the algorithm. We choose here to terminate when a maximum number of iterations $K$ has been reached,
and terminate the algorithm earlier when the following condition is satisfied:
\begin{align*}
\min \left\{ | (\omega^{k+1} - \omega^k)_{i,j} |, \left| \frac{(\omega^{k+1} - \omega^k)_{i,j}}{(\omega^k)_{i,j}} \right| \right\}\leq \delta,
\end{align*}
where the matrix $\mathbf{\Omega^k}= (\omega^k)_{i,j}$ is the row concatenation of $\mathbf{U}$ and $\mathbf{v^T}$, and $\delta$ is a small positive number. This condition detects the convergence of the learning algorithm, and is verified whenever the change in $\mathbf{U}$ and $\mathbf{v}$ is very small. This termination criterion is used for example in \citet{sriperumbudur2007sparse}.


\begin{algorithm}[t]
\begin{algorithmic}
\STATE $\mathbf{1.}$ Choose any initial point: $\mathbf{U^0}$ and $\mathbf{v^0} \geq \epsilon$.
\STATE $\mathbf{2.}$ For $k=0, \dots, {K-1}$,
\STATE \hspace{2mm}$\mathbf{2.1}$ Compute $(\mathbf{A}, \mathbf{b}) = \nabla h (\mathbf{U^k}, \mathbf{v^k})$.
\STATE \hspace{2mm}$\mathbf{2.2}$ Solve with Algorithm \ref{alg:algo_sgd} the convex optimization problem:
\begin{align*}
(\mathbf{U^{k+1}}, \mathbf{v^{k+1}}) \leftarrow & \argmin_{(\mathbf{U}, \mathbf{v})} \{ g(\mathbf{U}, \mathbf{v}) - Tr(\mathbf{U}^T \mathbf{A}) - \mathbf{v}^T \mathbf{b} \} \\
															& \text{ subject to } \mathbf{v} \geq \epsilon.
\end{align*}
\STATE \hspace{2mm}$\mathbf{2.3}$ If $(\mathbf{U^{k+1}}, \mathbf{v^{k+1}}) \approx (\mathbf{U^{k}}, \mathbf{v^{k}})$, return $(\mathbf{U^{k+1}}, \mathbf{v^{k+1}})$.
\end{algorithmic}
\caption{\label{alg:algo_DCA}LAST (Learning Algorithm for Soft-Thresholding classifier)}
\end{algorithm}

\section{Experimental results}
\label{sec:exp_results}

In this section, we evaluate the performance of our classification algorithm on textures, digits and natural images datasets, and compare it to different competitor schemes. We expose in Section \ref{sec:parameter_selection} the choice of the parameters of the model and the algorithm. We then focus on the experimental assessment of our scheme. Following the methodology of  \citet{coates2011importance}, we break the feature extraction algorithms into (i) a learning algorithm (e.g, K-Means) where a set of basis functions (or dictionary) is learned and (ii) an encoding function (e.g., $\ell_1$ sparse coding) that maps an input point to its feature vector. In a first step of our analysis (Section \ref{sec:sec_exp1}), we therefore \textit{fix the encoder} to be the soft-thresholding mapping and compare LAST to existing supervised and unsupervised learning techniques. Then, in the following subsections, we compare our complete classification architecture (i.e., learning and encoding function) to several classifiers, in terms of accuracy and efficiency. In particular, we show that our proposed approach is able to compete with recent classifiers, despite its simplicity.

\subsection{Parameter selection} 
\label{sec:parameter_selection}

We first discuss the choice of the model parameters for our method. Unless stated otherwise, we choose the vector $\mathbf{s}$ according to the distribution of the different classes in the training set.
We set the value of the regularization parameter to $\nu = 1$, as it was found empirically to be a good choice in our experiments. It is worth mentioning that setting $\nu$ by cross-validation might give better results, but it would also be computationally more expensive. We set moreover the parameter of the soft-thresholding mapping approximation to $\beta = 100$. Recall finally that the sparsity parameter $\alpha$ is always equal to $1$ in our method, and therefore does not require any manual setting or cross-validation procedure. 

In all experiments, we have moreover chosen to initialize LAST by setting $\mathbf{U^0}$ equal to a random subsample of the training set, and $\mathbf{v^0}$ is set to the vector whose entries are all equal to $1$. We however noticed empirically that choosing a different initialization strategy does not significantly change the testing accuracy.
Then, we fix the maximum number of iterations of LAST to $K=50$. Moreover, setting properly the parameters $t_0$ and $\rho$ in Algorithm \ref{alg:algo_sgd} is quite crucial in controlling the convergence of the algorithm. In all the experiments, we have set the parameter $t_0 = T / 10$, where $T$ denotes the number of iterations. Furthermore, during the first $T / 20$ iterations, several values of $\rho$ are tested $\{  0.1, 0.01, 0.001 \}$, and the value that leads to the smallest objective function is chosen for the rest of the iterations. Finally, the minibatch size in Algorithm \ref{alg:algo_sgd} depends on the size of the training data. In particular, when the size of the training data $m$ is relatively small (i.e., smaller than $5000$), we used a batch gradient descent, as the computation of the (complete) gradient is tractable. In this case, we set the number of iterations to $T = 1000$. Otherwise, we use a batch size of $200$, and perform $T = 5000$ iterations of the stochastic gradient descent in Algorithm \ref{alg:algo_sgd}.

\subsection{Analysis of the learning algorithm}
\label{sec:sec_exp1}
In a first set of experiments, we focus on the comparison of our learning algorithm (LAST) to other learning techniques, and fix the encoder to be the soft-thresholding mapping for all the methods. 
We present a comparative study on textures and natural images classification tasks.

\subsubsection{Experimental settings}
\label{sec:sec_exp1_settings}
We consider the following dictionary learning algorithms:
\begin{enumerate}
\item \textbf{Supervised random samples:} The atoms of $\mathbf{D}$ are chosen randomly from the training set, in a supervised manner. That is, if $\kappa$ denotes the desired proportion of class `$1$' atoms in the dictionary, the dictionary is built by randomly picking $\kappa N$ training samples from class `$1$' and $(1- \kappa) N$ samples from class `$-1$', where $N$ is the number of atoms in the dictionary. 
\item \textbf{Supervised K-means:} We build the dictionary by merging the subdictionaries obtained by applying the K-means algorithm successively to training samples of class `$1$' and `$-1$', 
where the number of clusters is fixed respectively to $\kappa N$ and $(1-\kappa) N$.
\item \textbf{Dictionary learning for } $\mathbf{\ell_1}$ \textbf{ sparse coding:} The dictionary $\mathbf{D}$ is built by solving the classical dictionary learning problem for $\ell_1$ sparse coding:
\begin{align}
\min_{\mathbf{D}, \mathbf{c_i}} \sum_{i=1}^m \| \mathbf{x_i} - \mathbf{D} \mathbf{c_i} \|_2^2 + \lambda \| \mathbf{c_i} \|_1 \text{ subject to } \forall j, \| \mathbf{d_j} \|_2^2 \leq 1.
\label{eq:dictlearning_l1}
\end{align}
To solve this optimization problem, we used the algorithm proposed by \citet{mairal2010online} and implemented in the SPAMS package. The parameter $\lambda$ is chosen by a cross-validation procedure in the set $\{ 0.1, 0.01, 0.001 \}$. Note that, while the previous two learning algorithms make use of the labels, this algorithm is unsupervised.
\item \textbf{Stochastic Gradient Descent (SGD):} The dictionary $\mathbf{D}$ and classifier $\mathbf{w}$ are obtained by optimizing the following objective function using \textit{mini-batch stochastic gradient descent}:
\begin{align*}
J(\mathbf{D}, \mathbf{w}) = \sum_{i=1}^m L (y_i \mathbf{w^T}  q(\mathbf{D^T x_i} - \alpha)) + \frac{\nu}{2} \| \mathbf{w} \|_2^2,
\end{align*}
with $q(x) = \frac{1}{\beta} \log( 1 + \exp( \beta x ) )$. This corresponds to the original objective function in Eq. (\ref{eq:prob_formulation}), where $h_{\alpha}$ is replaced with its smooth approximant.\footnote{We also tested SGD on the original (non-smooth) optimization problem. This resulted in slightly worse performance. We therefore only report results obtained on the smoothed objective function.} This smoothing procedure is similar to the one used in our relaxed formulation (Section \ref{sec:relaxed_formulation}). As in LAST, we set $\beta = 100$, $\alpha = 1$, and use the same initialization strategy. This setting allows us to directly compare LAST and this generic stochastic gradient descent procedure widely used for training neural networks.
Following \citet{glorot2011deep}, we use a mini-batch size of $10$, and use a constant step size chosen in $\{0.1, 0.01, 0.001, 0.0001 \}$. The stepsize is chosen through a cross-validation procedure, with a randomly chosen validation set made up of $10\%$ of the training data. The number of iterations of SGD is set to $250000$.
\end{enumerate}

For the first three algorithms, the parameter $\alpha$ in the soft-thresholding mapping is chosen with cross validation in $\{ 0.1, 0.2, \dots, 0.9, 1 \}$. The features are then computed by applying the soft thresholding map $h_{\alpha}$, and a linear SVM classifier is trained in the feature space. For the random samples and $K$-means approaches, we set $\kappa = 0.5$ as we consider classification tasks with roughly equal number of training samples from each class. 
Finally, for SGD and LAST, the dictionary $\mathbf{D}$ and linear classifier $\mathbf{w}$ are learned simultaneously. The encoder $h_{1}$ is used to compute the features. 

\subsubsection{Experimental results}
\label{sec:sec_exp1_results}
\begin{figure}[ht]
\centering
\includegraphics[width=0.4\textwidth]{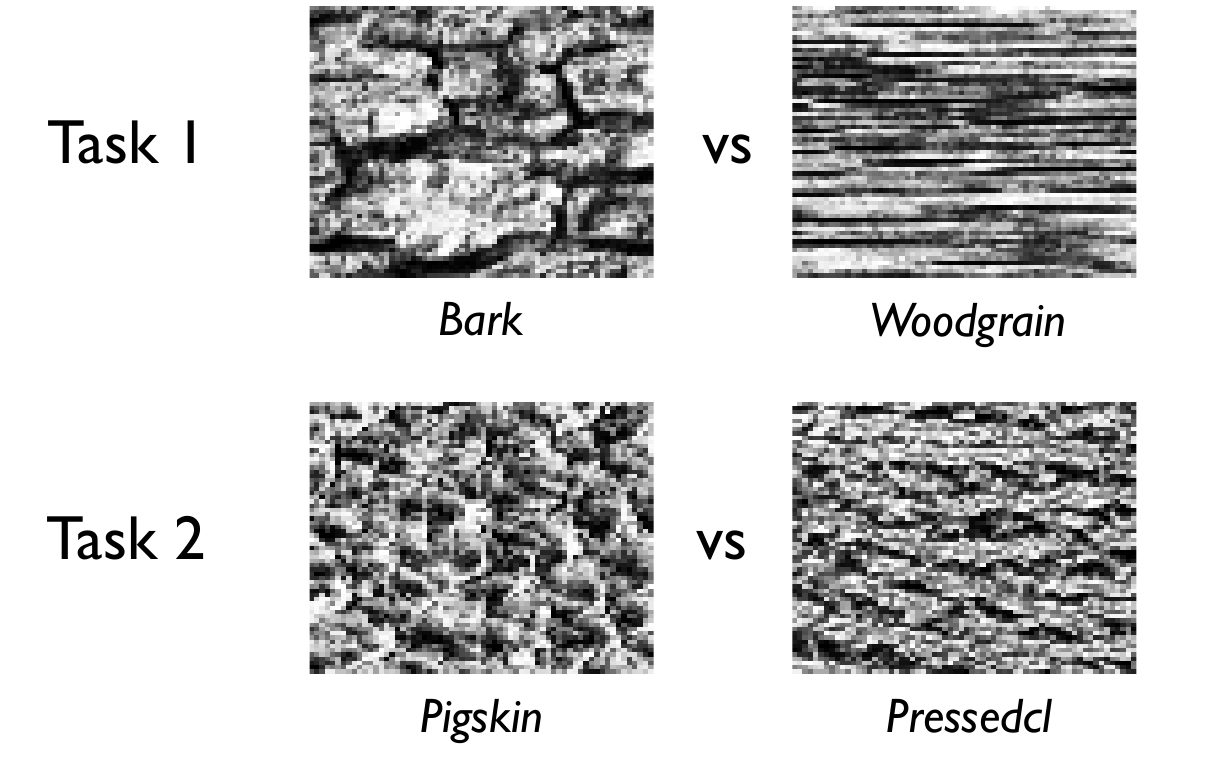}
\caption{\label{fig:texture_images}Two binary classification tasks (\textit{bark} vs \textit{woodgrain} and \textit{pigskin} vs. \textit{pressedcl})}
\end{figure}

\begin{figure}[ht]
\centering
\subfigure[\textit{Bark vs. Woodgrain}]{
\includegraphics[width=0.4\textwidth]{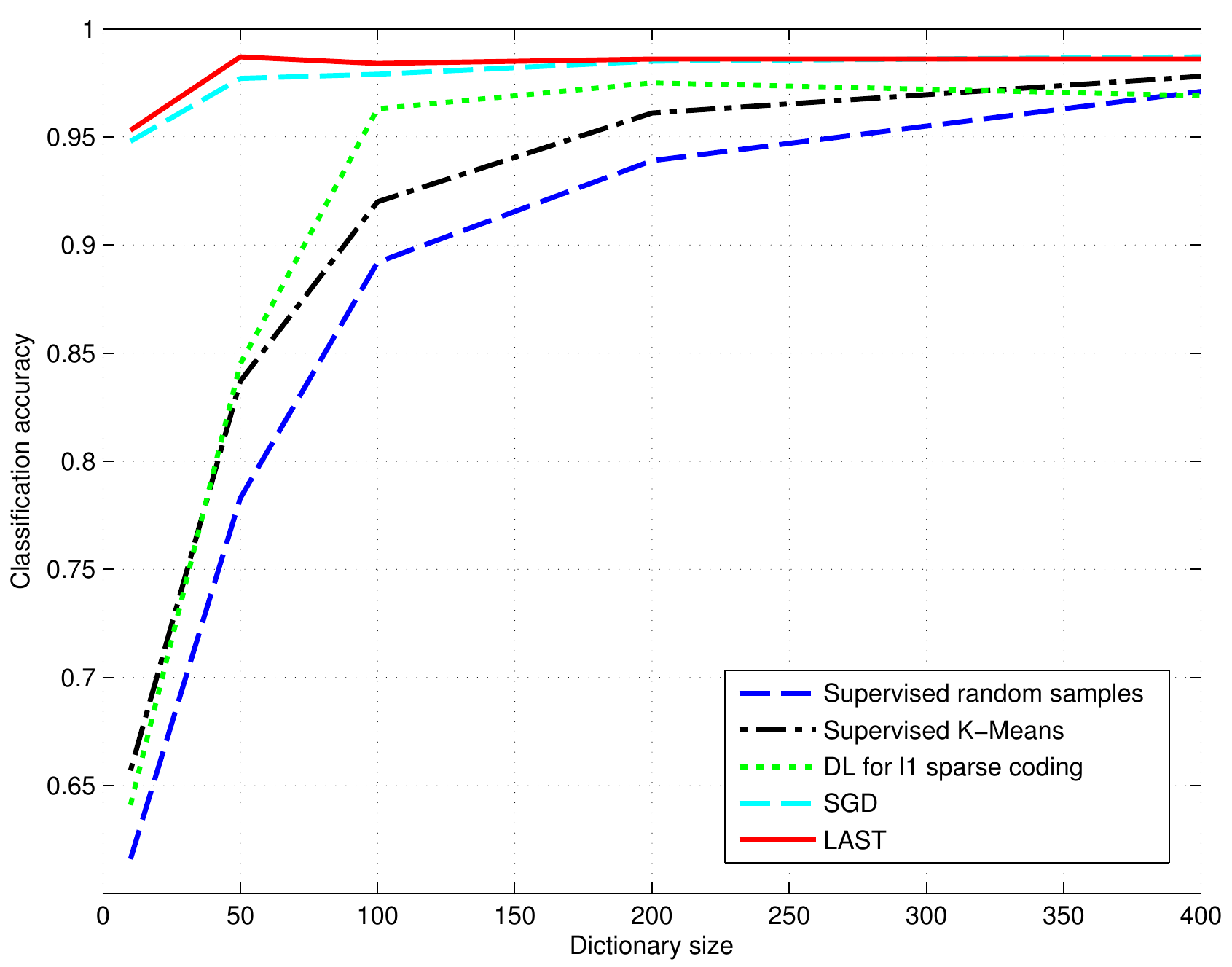}
}
\subfigure[\textit{Pigskin vs. Pressedcl}]{
\includegraphics[width=0.4\textwidth]{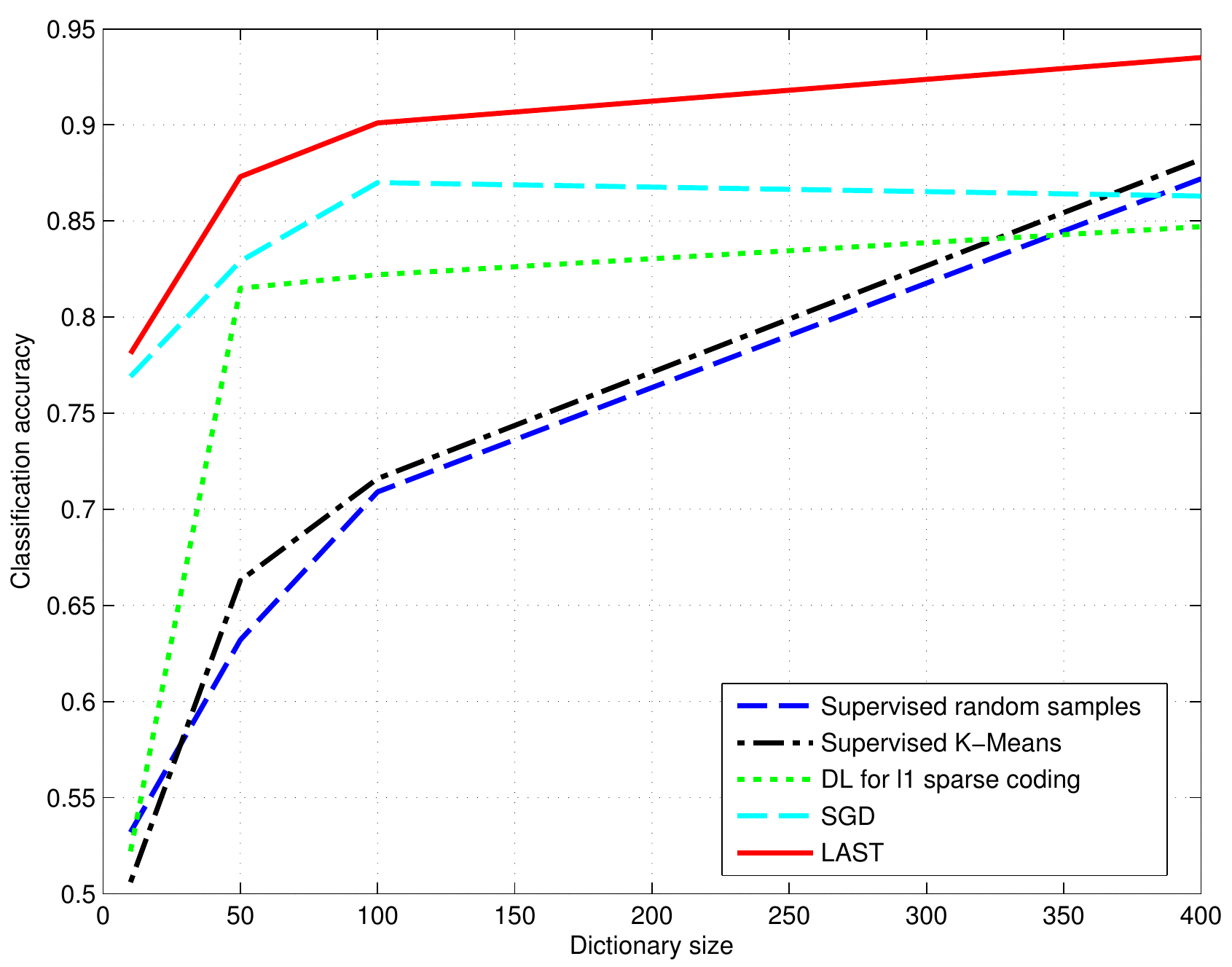}
}
\caption{\label{fig:results_dl_textures}Texture classification results (fixed soft-thresholding encoder)}
\end{figure}

In our first experiment, we consider two binary texture classification tasks, where the textures are collected from the 32 Brodatz dataset \citep{valkealahti1998reduced} and shown in Fig. \ref{fig:texture_images}. For each pair of textures under test, we build the training set by randomly selecting $500$ $12 \times 12$ patches per texture, and the test data is constructed similarly by taking $500$ patches per texture. The test data does not contain any of the training patches. All the patches are moreover normalized to have unit $\ell_2$ norm. Fig. \ref{fig:results_dl_textures} shows the binary classification accuracy of the soft-thresholding based classifier as a function of the dictionary size, for dictionaries learned with the different algorithms. 

For the first task (\textit{bark} vs. \textit{woodgrain}), one can see that LAST and SGD dictionary learning methods outperform the other methods for small dictionary sizes. For large dictionaries (i.e., $N \approx 400$) however, all the learning algorithms yield approximately the same classification accuracy. This result is in agreement with the conclusions of \citet{coates2011importance}, where the authors show empirically that the choice of the learning algorithm becomes less crucial when dictionaries are very large.
In the second and more difficult classification task (\textit{pigskin} vs. \textit{pressedcl}), our algorithm yields the best classification accuracy for all tested dictionary sizes ($10 \leq N \leq 400$). Interestingly, unlike the previous task, the design of the dictionary is crucial for all tested dictionary sizes. Using much larger dictionaries might result in performance that is close to the one obtained using our algorithm, but comes at the price of additional computational and memory costs.

\begin{figure}[ht]
\centering
\includegraphics[width=0.4\textwidth]{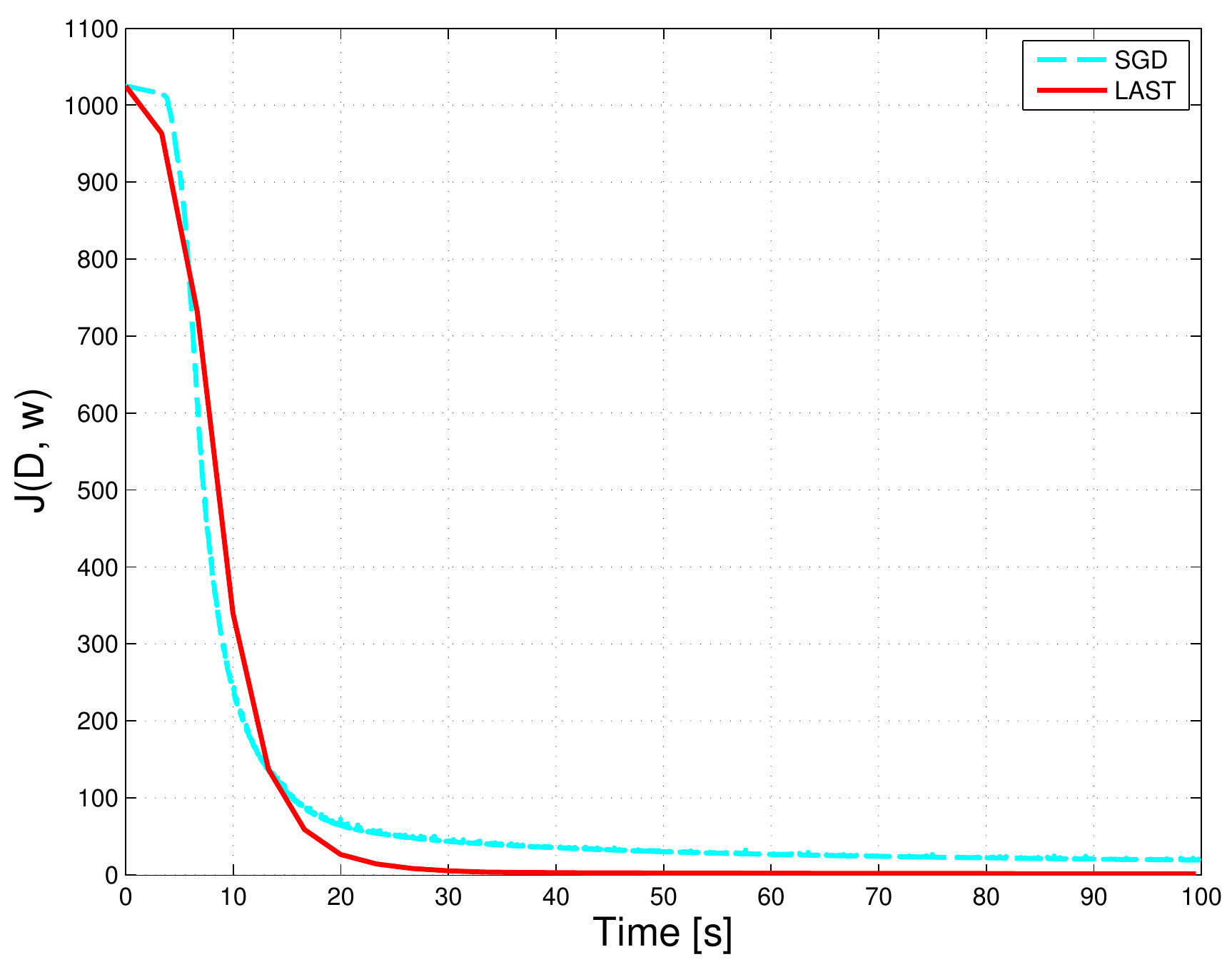}
\caption{$J(\mathbf{D}, \mathbf{w})$ as a function of the elapsed time [s] for Stochastic Gradient Descent and LAST. For SGD: $J(\mathbf{D}_{t=100}, \mathbf{w}_{t=100}) = 19$, LAST: $J(\mathbf{D}_{t=100}, \mathbf{w}_{t=100}) = 1.4$.\label{fig:iters_LAST_SGD_GD}}
\end{figure}

Fig. \ref{fig:iters_LAST_SGD_GD} further illustrates the evolution of the objective function with respect to the elapsed training time for LAST and SGD, for a dictionary of size $50$. 
One can see that LAST quickly converges to a solution with a small objective function. On the other hand, SGD reaches a solution with larger objective function than LAST. 



\begin{figure}[h!]
\centering
\includegraphics[width=0.5\textwidth]{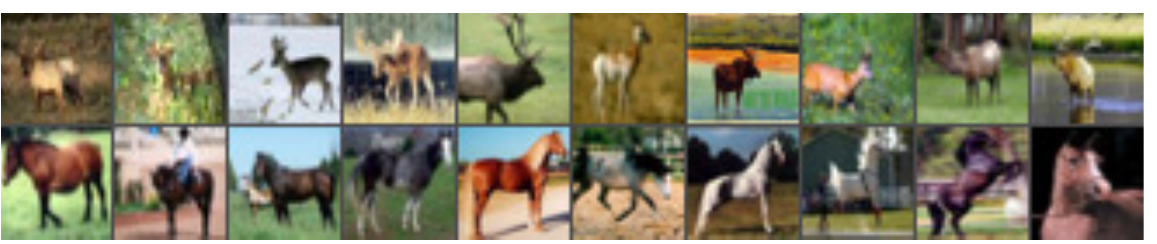}
\caption{\label{fig:cifar_training_images} Examples of CIFAR-10 images in categories ``deer'' and ``horse''.}
\end{figure}

\begin{figure}[h!]
\centering
\includegraphics[width=0.4\textwidth]{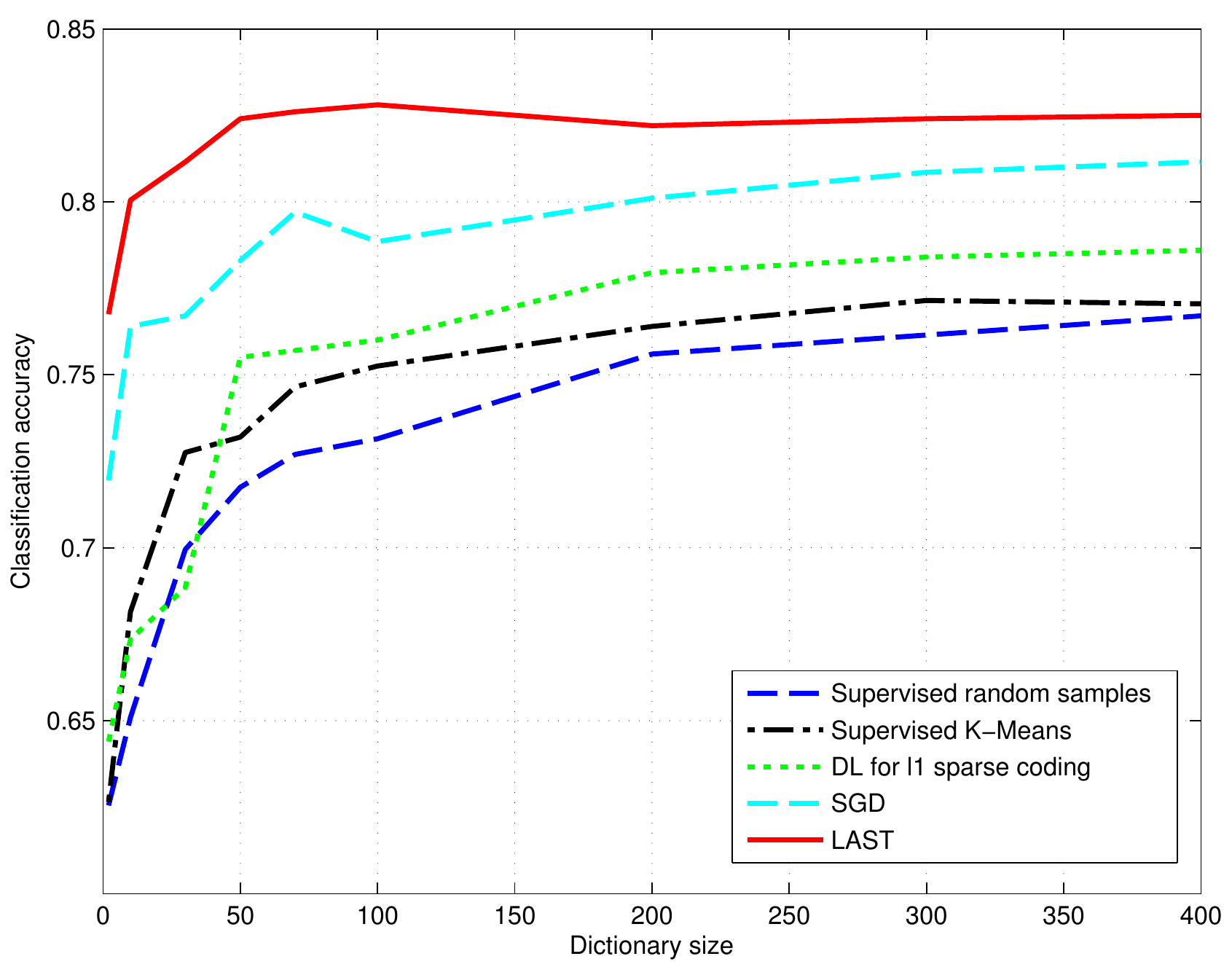}
\caption{\label{fig:cifar_bin_fixedEncoder}  Performance of the ``deer'' vs. ``horse'' binary classification task (fixed soft-thresholding encoder)}
\end{figure}

We now conduct experiments on the popular CIFAR-10 image database \citep{krizhevsky2009learning}. The dataset contains $10$ classes of $32 \times 32$ RGB images. For simplicity and better comparison of the different learning algorithms, we restrict in a first stage the dataset to the two classes ``deer'' and ``horse''. We extend our results to the multi-class scenario later in Section \ref{sec:cifar10_exp}. Fig. \ref{fig:cifar_training_images} illustrates some training examples from the two classes. The classification results are reported in Fig. \ref{fig:cifar_bin_fixedEncoder}. 

\begin{figure}[h!]
\centering
\subfigure[K-Means]{
\includegraphics[width=0.20\textwidth]{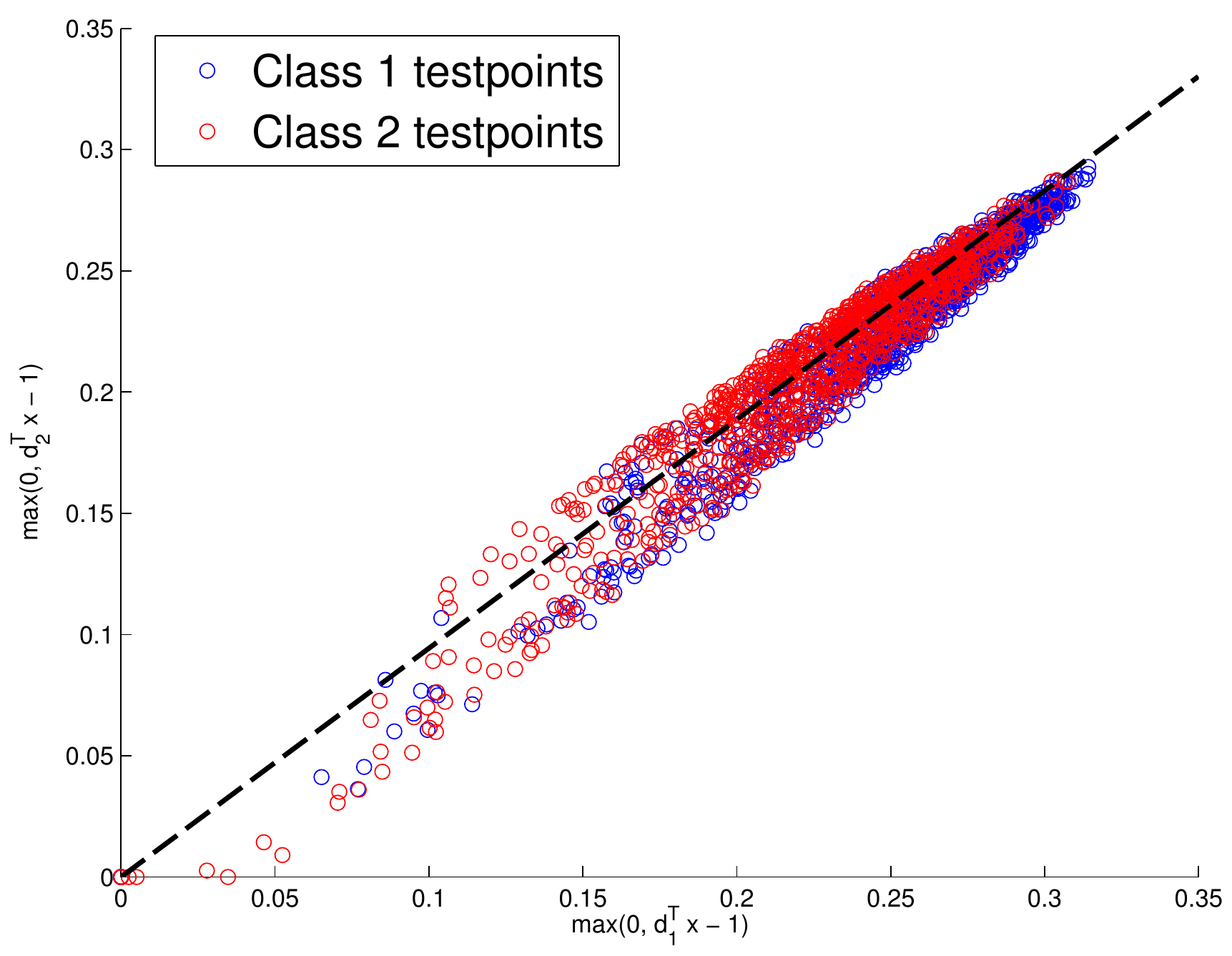}
}
\subfigure[LAST]{
\includegraphics[width=0.20\textwidth]{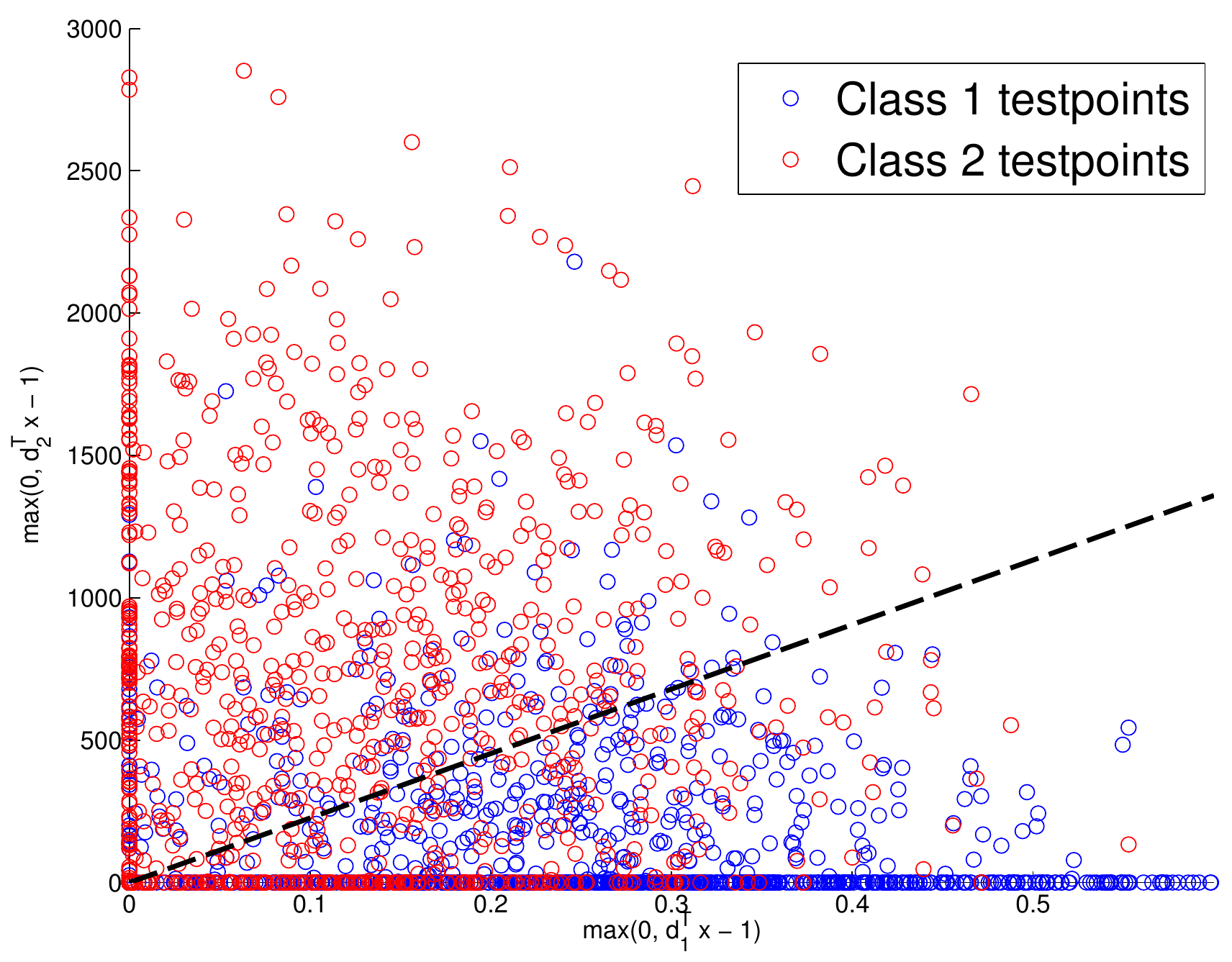}
}
\caption{\label{fig:2d_features}Learned 2D features and linear classifiers with K-Means and LAST for the ``deer'' vs. ``horse'' classification task ($N = 2$).}
\end{figure}

\begin{figure}[h!]
\centering
\subfigure[K-Means]{
\includegraphics[width=0.4\textwidth]{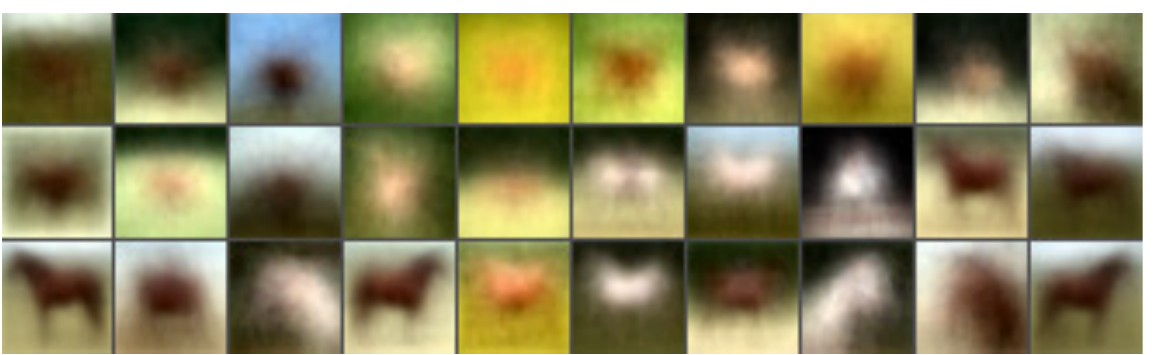}
}
\subfigure[LAST]{
\includegraphics[width=0.4\textwidth]{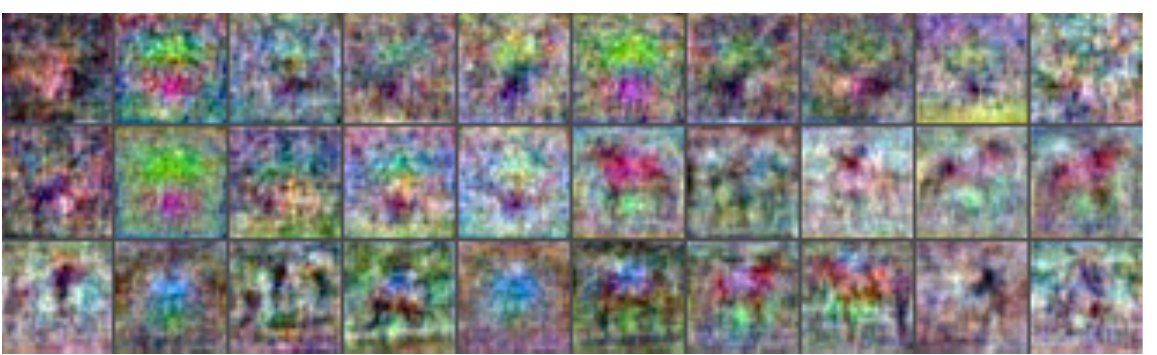}
}
\caption{\label{fig:dict_learned_LAST} Normalized dictionary atoms learned with K-Means and LAST, for the ``deer'' vs. ``horse'' binary classification task ($N = 30$).}
\end{figure}

Once again, the soft-thresholding based classifier with a dictionary and linear classifier learned with LAST outperforms all other learning techniques. In particular, using the LAST dictionary learning strategy results in significantly higher performance than stochastic gradient descent for all dictionary sizes. We further note that with a very small dictionary (i.e., $N = 2$), LAST reaches an accuracy of $77\%$, whereas some learning algorithms (e.g., K-means) do not reach this accuracy even with a dictionary that contains as many as $400$ atoms. To further illustrate this point, we show in Fig. \ref{fig:2d_features} the 2-D testing features obtained with a dictionary of two atoms, when $\mathbf{D}$ is learned respectively with the K-Means method and LAST. Despite the very low-dimensionality of the feature vectors, the two classes can be separated with a reasonable accuracy using our algorithm (Fig. \ref{fig:2d_features} (b)), whereas features obtained with the K-means algorithm clearly cannot be discriminated (Fig. \ref{fig:2d_features} (a)). 
We finally illustrate in Fig. \ref{fig:dict_learned_LAST} the dictionaries learned using K-Means and LAST for $N = 30$ atoms. It can be observed that, while K-Means dictionary consists of smoothed images that minimize the reconstruction error, our algorithm learns a discriminative dictionary whose goal is to underline the difference between the images of the two classes. 

In summary, our supervised learning algorithm, specifically tailored for the soft-thresholding encoder provides significant improvements over traditional dictionary learning schemes. Our classifier can reach high accuracy rates, even with very small dictionaries, which is not possible with other learning schemes.

\subsection{Classification performance on binary datasets}
\label{sec:sec_exp2}

In this section, we compare the proposed LAST classification method\footnote{By extension, we define the LAST classifier to be the soft-thresholding based classifier, where the parameters $(\mathbf{D}, \mathbf{w})$ are learned with LAST.} to other classifiers.
Before going through the experimental results, we first present the different methods under comparison:
\begin{enumerate}
\item \textbf{Linear SVM: } We use the efficient Liblinear \citep{REF08a} implementation for training the linear classifier. The regularization parameter is chosen using a cross-validation procedure.
\item \textbf{RBF kernel SVM: } We use LibSVM \citep{CC01a} for training. Similarly, the regularization and width parameters are set with cross-validation.
\item \textbf{Sparse coding:} Similarly to the previous section, we train the dictionary by solving Eq. (\ref{eq:dictlearning_l1}). We use however the encoder that ``matches naturally" with this training algorithm, that is:
\begin{align*}
\argmin_{\mathbf{c}} \| \mathbf{x} - \mathbf{D c} \|_2^2 + \lambda \| \mathbf{c} \|_1,
\end{align*}
where $\mathbf{x}$ is the test sample, $\mathbf{D}$ the previously learned dictionary and $\mathbf{c}$ the resulting feature vector. A linear SVM is then trained on the resulting feature vectors. This classification architecture, denoted ``sparse coding'' below, is similar to that of \citet{raina2007self}.
\item \textbf{Nearest neighbor classifier (NN):} Our last comparative scheme is a nearest neighbor classifier where the dictionary is learned using the supervised K-means procedure described in \ref{sec:sec_exp1_settings}. At test time, the sample is assigned the label of the dictionary atom (i.e., cluster) that is closest to it.
\end{enumerate}

Note that we have dropped the supervised random samples learning algorithm used in the previous section as it was shown to have worse classification accuracy than the K-means approach. 


\begin{table}[ht]
\centering
\begin{tabular}{|l||c|c|}
\hline
& \multicolumn{1}{|c|}{Task 1 [\%]} & \multicolumn{1}{|c|}{Task 2 [\%]}  \\
 \hline
Linear SVM & 49.5 & 49.1 \\  
  \hline
RBF kernel SVM & 98.5 & 90.1\\  
  \hline
Sparse coding ($N = 50$) & 97.5 & 85.5 \\
  \hline
Sparse coding ($N = 400$) & 98.1 & 90.9 \\
  \hline
NN ($N = 50$) & 94.3 & 84.1 \\
\hline
NN ($N = 400$) & 97.8 & 86.6 \\
\hline
LAST ($N = 50$) & \textbf{98.7} & 87.3  \\
\hline
LAST ($N = 400$) & 98.6 & \textbf{93.5} \\
\hline
\end{tabular}
\caption{\label{tab:classification_accuracy} Classification accuracy for binary texture classification tasks.}
\footnotesize
\end{table}

Table \ref{tab:classification_accuracy} first shows the accuracies of the different classifiers in the two binary textures classification tasks described in \ref{sec:sec_exp1_results}. In both experiments, the linear SVM classifier results in a very poor performance, which is close to the random classifier. This suggests that the considered task is nonlinear, and has to be tackled with a nonlinear classifier. One can see that the RBF kernel SVM results in a significant increase in the classification accuracy. Similarly, the $\ell_1$ sparse coding non linear mapping also results in much better performance compared to the linear classifier, while the nearest neighbor approach performs a bit worse than sparse coding. We note that, for a fixed dictionary size, our classifier outperforms NN and sparse coding classifiers in both tasks. Moreover, it provides comparable or superior performance to the RBF kernel SVM in both tasks. 

We now turn to the binary experiment ``deer'' vs. ``horse'' described in the previous subsection. We show the classification accuracies of the different classifiers in Table \ref{tab:classification_accuracy_cifarbin}. LAST outperforms sparse coding and nearest neighbour classifiers for the tested dictionary sizes. RBF kernel SVM however slightly outperforms LAST with $N = 100$ in this experiment. Note however that the RBF kernel SVM approach is much slower at test time, which makes it impractical for large-scale problems.


\begin{table}[ht]
\centering
\begin{tabular}{|l||c|}
\hline
& \multicolumn{1}{|c|}{``deer'' vs. ``horse'' [\%]} \\
 \hline
Linear SVM & 72.6 \\
  \hline
RBF kernel SVM & \textbf{83.5} \\  
  \hline
Sparse coding ($N = 10$) & 70.6 \\
  \hline
Sparse coding ($N = 100$) &  76.2 \\
\hline
NN ($N = 10$) & 67.7 \\
\hline
NN ($N = 100$) & 70.9 \\
\hline
LAST ($N = 10$) & 80.1 \\
\hline
LAST ($N = 100$) & 82.8 \\
\hline
\end{tabular}
\caption{\label{tab:classification_accuracy_cifarbin} Binary classification accuracy on the binary classification problem ``deer'' vs. ``horse''.}
\footnotesize
\end{table}

Overall, the proposed LAST classifier compares favorably to the different tested classifiers. In particular, LAST outperforms the sparse coding technique for a fixed dictionary size in our experiments. This result is notable, as sparse coding classifiers are known to provide very good classification performance in vision tasks. Note that, when used with another standard learning approach as K-Means, the soft-thresholding based classifier is outperformed by sparse coding, which shows the importance of the learning scheme in the success of this classifier.

\subsection{Handwritten digits classification}
We now consider a classification task on the MNIST \citep{lecun1998gradient} and USPS   \citep{hull1994database} handwritten digits datasets. USPS contains $9298$ images of size $16 \times 16$ pixels, with $7291$ images used for training and $2007$ for testing. The larger MNIST database is composed of $60 000$ training images and $10 000$ test images, all of size $28 \times 28$ pixels.  We preprocess all the images to have zero-mean and to be of unit Euclidean norm. 

We address the multi-class classification task using a one-vs-all strategy, as it is often done in classification problems. Specifically, we learn a separate dictionary and a binary linear classifier by solving the optimization problem for each one-vs-all problem. Classification is then done by predicting using each binary classifier, and choosing the prediction with highest score. In LAST, for each one-vs-all task, we naturally set $1/10$ of the entries of $\mathbf{s}$ to $1$ and the other entries to $-1$, assuming the distribution of features of the different classes in the dictionary should roughly be that of the images in the training set. 
In our proposed approach and SGD, we used dictionaries of size $N=200$ for USPS and $N=400$ for MNIST as the latter dataset contains much more training samples. 
\begin{table}[t]
\centering
\begin{tabular}{|l||c|c|}
\hline
 & \multicolumn{1}{|c|}{MNIST}  & \multicolumn{1}{|c|}{USPS}\\
\hline
Linear SVM & 8.19 & 9.07 \\
 \hline
RBF kernel SVM & 1.4 & 4.2 \\  
  \hline
K-NN $\ell_2$ & 5.0 & 5.2 \\
  \hline
LAST & 1.32 & 4.53 \\
 \hline
Sparse coding & 3.0 & 5.33 \\
 \hline
\citet{huang2006sparse} & - & 6.05 \\
\hline
SDL-G L \citep{mairal2008supervised} & 3.56 & 6.67 \\
  \hline
SDL-D L \citep{mairal2008supervised}  & 1.05 & 3.54\\
  \hline
\citet{ramirez2010classification} & 1.26 & 3.98 \\
\hline
SGD & 2.22 & 5.88 \\
\hline
3 layers ReLU net \citep{glorot2011deep} & 1.43 & - \\
\hline
\end{tabular}
\caption{\label{tab:digits_classification_accuracy} Classification error (percentage) on MNIST and USPS datasets.}
\footnotesize
\end{table}
We compare LAST to baseline classification techniques described in the previous section, as well as to sparse coding based methods. 
In addition to building the dictionary in an unsupervised way, we consider the sparse coding classifiers in \citet{mairal2008supervised, huang2006sparse, ramirez2010classification}, which construct the dictionary in a supervised fashion.

Classification results are shown in Table \ref{tab:digits_classification_accuracy}. 
One can see that LAST largely outperforms linear and nearest neighbour classifiers. Moreover, our method has a slightly better accuracy than RBF-SVM in MNIST, while being slightly worse on the USPS dataset. Our approach also outperforms the soft-thresholding based classifier optimized with stochastic gradient descent on both tasks, which highlights the benefits of our optimization technique compared to the standard algorithm used for training neural networks. We also report from \citet{glorot2011deep} the performance of a three hidden layer rectified network optimized with stochastic gradient decent, without unsupervised pre-training. It can be seen that LAST, while having a much simpler architecture, slightly outperforms the deep rectifier network on the MNIST task.
Furthermore, LAST outperforms the unsupervised sparse coding classifier in both datasets. Interestingly, the proposed scheme also competes with, and sometimes outperforms the discriminative sparse coding techniques of \citep{huang2006sparse, mairal2008supervised, ramirez2010classification}, where the dictionary is tuned for classification. While providing comparable results, the LAST classifier is much faster at test time than sparse coding techniques and RBF-SVM classifiers. It is noteworthy to mention that the best discriminative dictionary learning results we are aware of on these datasets are achieved by \citet{mairal2012task} with an error rate of $0.54\%$ on MNIST and $2.84\%$ on USPS. Note however that in this paper, the authors explicitly incorporate translation invariance in the problem by augmenting the training set with shifted versions of the digits. Our focus goes here instead on methods that do not augment the training set with distorted or transformed samples. 

\subsection{CIFAR-10 classification}
\label{sec:cifar10_exp}

We now consider the multi-class classification problem on the CIFAR-10 dataset \citep{krizhevsky2009learning}. The dataset contains $60000$ color images of size $32 \times 32$ pixels, with $50000$ images for training and $10000$ for testing. The classifier input consists of vectors of raw pixel values of dimension $32 \times 32 \times 3 = 3072$. This setting, similar to that of \citet{glorot2011deep}, takes no advantage of the fact that we are dealing with images and is sometimes referred to as ``permutation invariant'', as columns in the data could be shuffled without affecting the result. We consider this scenario to focus on the comparison of the performance of the classifiers. 
Due to the relatively high dimensions of the problem ($n = 3072$, $m = 50000$), we limit ourselves to  classifiers with feedforward architectures. In fact, using RBF-SVM for this task would be prohibitively slow at the training and testing stage. For each one-vs-all task, we set the dictionary size of LAST and SGD methods to $400$. Moreover, unlike the previous experiment, we set in LAST half of the entries of the sign vector $\mathbf{s}$ to $1$ and the other half to $-1$. This is due to the high variability of intra-class images and the relatively small dictionary size: the number of atoms required to encode the positive class might not be sufficient if $\mathbf{s}$ is set according to the distribution of images in the training set. The results are reported in Table \ref{tab:cifar_classification_accuracy}.

\begin{table}[t]
\centering
\begin{tabular}{|l||c|}
\hline
 & \multicolumn{1}{|c|}{CIFAR-10}\\
\hline
Linear SVM & 59.70 \\
  \hline
LAST $(N = 400)$ & 46.56 \\
\hline
SGD $(N = 400)$ & 52.96 \\
 \hline
3 layers ReLU net  &  50.86 \\
\hline
3 layers ReLU net + sup. pre-train &  49.96 \\
\hline
\end{tabular}
\caption{\label{tab:cifar_classification_accuracy} Classification error (percentage) on the CIFAR-10 dataset. ReLU net results are reported from \citep{glorot2011deep}.}
\footnotesize
\end{table}

Once again, this experiment confirms the superiority of our learning algorithm over linear SVM. Moreover, LAST significantly outperforms the generic SGD training algorithm (by more than $6 \%$) in this challenging classification example. What is more surprising is that LAST significantly surpasses the rectifier neural network with $3$ hidden layers \citep{glorot2011deep} trained using a generic stochastic gradient descent algorithm (with or without pre-training). This shows that, despite the simplicity of our architecture (it can be seen as one hidden layer), the adequate training of the classification scheme can give better performance than complicated structures that are potentially difficult to train. We finally report the results of sparse coding classifier with a dictionary trained using Eq. (\ref{eq:dictlearning_l1}). If we use a dictionary with $400$ atoms, we get an error of $53.9\%$. By using a much larger dictionary of $4000$ atoms, the error reduces to  $46.5\%$. The computation of the test features is however computationally very expensive in that case.
\section{Discussion}
\label{sec:discussion}
We first discuss in this section aspects related to the computational complexity of LAST. Then, we analyze the sparsity of the obtained solutions. We finally explain some of the differences between LAST and the generic stochastic gradient descent algorithm. 
\subsection{Computational complexity at test time}
We compare the computational complexity and running times of LAST classifier to the ones of different classification algorithms. 
Table \ref{tab:complexity} shows the computational complexity for classifying one test sample using various classifiers and the time needed to classify MNIST test images. 
We recall that $n$, $m$, and $N$ denote respectively the signals dimension, the number of training samples and the dictionary size. Clearly, linear classification is very efficient as it only requires the computation of one inner product between two vectors of dimension $n$. Nonlinear SVMs however have a test complexity that is linear in the number of support vectors, which scales linearly with the training size \citep{burges1998tutorial}. This solution is therefore not practical for relatively large training sets, like MNIST or CIFAR-10. Feature extraction with sparse coding involves solving an optimization problem, which roughly requires $1 / \sqrt{\epsilon}$ matrix-vector multiplications, where $\epsilon$ controls the precision \citep{beck2009fast}. For a typical value of $\epsilon = 10^{-6}$, the complexity becomes $1000 n N$ (neglecting other constants), that is $3$ orders of magnitude larger than the complexity of the proposed method. This can be seen clearly in the computation times, as our approach is slightly more expensive than linear SVM, but remains much faster than other methods. Note moreover that the soft-thresholding classification scheme is very simple to implement in practice at test time, as it is a direct map that only involves $\max$ and linear operations.
\subsection{Sparsity}
\label{sec:exp_sparsity}
Sparsity is a highly beneficial property in representation learning, as it helps decomposing the factors of variations in the data into high level features \citep{bengio2013representation, glorot2011deep}. To assess the sparsity of the learned representation, we compute the average sparsity of our representation over all data points (training and testing combined) on the MNIST and CIFAR-10 dataset. We obtain an average of $96.7\%$ zeros in the MNIST case, and $95.3\%$ for CIFAR-10. In other words, our representations are very sparse, without adding an explicit sparsity penalization as in \citep{glorot2011deep}. Interestingly, the reported average sparsity in \citep{glorot2011deep} is $83.4\%$ on MNIST and $72.0\%$ on CIFAR-10. Our one-layer representation therefore exhibits an interesting sparsity property, while providing good predictive performance. 
\begin{table}[t]
\footnotesize
\centering
\begin{tabular}{|l|c|c|}
\hline
& Complexity & Time [s] \\
\hline
Linear SVM & $O(n)$ &  0.4 \\
\hline
RBF kernel SVM & $O(nm)$ & 154 \\
\hline
Sparse coding & $O\left( \frac{n N}{\sqrt{\epsilon}} \right)$\tablefootnote{The complexity reported here is that of the FISTA algorithm \cite{beck2009fast}, where $\epsilon$ denotes the required precision. Note that another popular method for solving sparse coding is the homotopy method, which is efficient in practice, however it has exponential theoretical complexity \cite{mairal2012complexity}.} & 14 \tablefootnote{To provide a fair comparison with our method, we used dictionaries of the same size as for our proposed approach, for the sake of this experiment.}  \\
\hline
LAST classifier & $O(n N)$ & 1.0 \\
\hline
\end{tabular}
\caption{\label{tab:complexity}Computational complexity for classifying one test sample, and time needed to predict the labels of the $10 000$ test samples in the MNIST dataset. For reference, all the experiments are carried out on a 2.6 GHz Intel Core i7 machine with 16 GB RAM.}
\end{table}
\subsection{LAST vs. stochastic gradient descent}

As discussed earlier, the soft-thresholding classification scheme belongs to the more general neural network models. Neural networks are commonly optimized with stochastic gradient descent algorithms, as opposed to the DC method proposed in this paper. 
The proposed learning algorithm has several advantages compared to SGD:
\begin{itemize}
\item \textbf{Better local minimum:} In all our experiments, LAST reached a better solution than SGD in terms of the testing accuracy. This confirms the observations of \citet{tao1998dc} whereby DCA converges to ``good'' local minima, and often to global minima in practice.
\item \textbf{Descent method:} Unlike stochastic gradient descent, LAST (and more generally DCA) is a descent method. Moreover, it is guaranteed to converge to a critical point \citep{tao1998dc}.
\item \textbf{No stepsize selection:} Stochastic gradient descent (and more generally gradient descent based algorithms) are very sensible to the difficult choice of the stepsize. Choosing a large stepsize in SGD can be beneficial as it helps escaping local minimas, but it can also lead to an oscillatory behaviour that prevents convergence. Interestingly, our optimization algorithm does not involve any stepsize selection, when given a convex optimization solver. In fact, our algorithm solves a sequence of convex problems, which can be solved with any off-the-shelf convex solver. Note that even if the intermediate convex optimization problems are solved with a gradient-descent based technique, the choice of the stepsize is less challenging as we have a better understanding of the theoretical properties of stepsize rules in convex optimization problems.
\end{itemize} 

As we have previously mentioned, unlike SGD, our algorithm assumes the sign vector of the linear classifier $\mathbf{w}$ to be known. A simple heuristic choice of this parameter was shown however to provide very good results in the experiments, compared to SGD. Of course, choosing this parameter with cross-validation might lead to better results, but also implies a slower training procedure.

\section{Conclusion}
\label{sec:conclusion}

We have proposed a supervised learning algorithm tailored for the soft thresholding based classifier. The learning problem, which jointly estimates a discriminative dictionary $\mathbf{D}$ and a classifier hyperplane $\mathbf{w}$ is cast as a DC problem and solved efficiently with an iterative algorithm. The proposed algorithm (LAST), which leverages the DC structure, significantly outperforms stochastic gradient descent in all our experiments. Furthermore, the resulting classifier consistently leads to better results than the unsupervised sparse coding classifier. Our method moreover compares favorably to other standard techniques as linear, RBF kernel or nearest neighbour classifiers. The proposed LAST classifier has also been shown to compete with recent discriminative sparse coding techniques in handwritten digits classification experiments.
We should mention that, while the sparse coding encoder features some form of competition between the different atoms in the dictionary (often referred to as \textit{explaining-away} \citep{gregor2010learning}), our encoder acts on the different atoms independently. Despite its simple behavior, our scheme is competitive when the dictionary and classifier parameters are learned in a suitable manner. 

The classification scheme adopted in this paper can be seen as a one hidden layer neural network with a soft-thresholding activation function. This activation function has recently gained significant attention in the deep learning community, as it is believed to make the training procedure easier and less prone to bad local minima. 
Our work reveals an interesting structure of the optimization problem for the one-hidden layer version of that network that allows to reach good minima.
An interesting question is whether it is possible to find a similar structure for networks with many hidden layers. This would help the training of deep networks, and offer insights on this challenging problem, which is usually tackled using stochastic gradient descent.

\appendix

\section{Soft-thresholding as an approximation to non-negative sparse coding}
\label{sec:softthresh_sparsecoding}

We show here that soft-thresholding can be viewed as a coarse approximation to the non-negative sparse coding mapping \citep{denil2012recklessly}. To see this, we consider the proximal gradient algorithm to solve the sparse coding problem with additional nonnegativity constraints on the coefficients. 
Specifically, we consider the following mapping
\begin{align*}
\argmin_{\mathbf{c} \in \mathbb{R}^N} \| \mathbf{x} - \mathbf{D c} \|_2^2 + \lambda \| \mathbf{c} \|_1 \text{ subject to } \mathbf{c} \geq \mathbf{0}.
\end{align*}
The proximal gradient algorithm proceeds by iterating the following recursive equation to convergence:   
\begin{align*}
\mathbf{c^{k+1}} = \text{prox}_{\lambda t \| \cdot \|_1 + \mathscr{I}_{\cdot \geq 0}} (\mathbf{c^k} + t \mathbf{D^T} ( \mathbf{x}  - \mathbf{D c^k} ) ),
\end{align*}
where $\text{prox}$ is the proximal operator, $t$ is the chosen stepsize and $\mathscr{I}_{\cdot \geq 0}$ is the indicator function, which is equal to $0$ if all the components of the vector are nonnegative, and $+\infty$ otherwise. Using the definition of the proximal mapping, we have 
\begin{align*}
\text{prox}_{\lambda t \| \cdot \|_1 + \mathscr{I}_{\cdot \geq 0}} (\mathbf{x}) & \triangleq \argmin_{\mathbf{u} \geq 0} \{ \frac{1}{2} \| \mathbf{u} - \mathbf{x} \|_2^2 + \lambda t \| \mathbf{u} \|_1 \} \\
																									  & = \max(0, \mathbf{x} - \lambda t).
\end{align*}
Therefore, imposing the initial condition $\mathbf{c^0} = \mathbf{0}$, and a stepsize $t = 1$, the first step of the proximal gradient algorithm can be written
\begin{align*}
\mathbf{c^1} = \max(0, \mathbf{D^T x} - \lambda) = h_{\lambda} (\mathbf{D^T x}),
\end{align*}
which precisely corresponds to our soft-thresholding map. In this way, our soft-thresholding map corresponds to an approximation of sparse coding, where only one iteration of proximal gradient algorithm is performed.
\section{Proofs}
\label{sec:proof_dc_decomp}

\subsection{Proof of Proposition \ref{prop:dc_nature}}

Before going through the proof of Proposition \ref{prop:dc_nature}, we need the following results in \cite[Section $4.2$]{horst2000introduction}:

\begin{proposition}
\label{prop:dc_properties}
\begin{enumerate}
\item Let $\{ f_i \}_{i=1}^l$ be DC functions. Then, for any set of real numbers $(\lambda_1, \dots, \lambda_l)$, $\sum_{i=1}^l \lambda_i f_i$ is also DC.
\item Let $f: \mathbb{R}^n \rightarrow \mathbb{R}$ be DC and $g: \mathbb{R} \rightarrow \mathbb{R}$ be convex. Then, the composition $g(f(\mathbf{x}))$ is DC.
\end{enumerate}
\end{proposition}

We recall that the objective function of (P) is given by:
\begin{align*}
\sum_{i=1}^m L \left( y_i \sum_{j=1}^N s_j q(\mathbf{u_j^T x_i} - v_j) \right) + \frac{\nu}{2} \| \mathbf{v} \|_2^2,
\end{align*}

The function $\| \mathbf{v} \|_2^2$ is convex and therefore DC. We show that the first part of the objective function is also DC. We rewrite this part as follows:
\begin{align*}
\sum_{i=1}^m L \left(\sum_{j: s_j = y_i} q(\mathbf{u_j^T x_i} - v_j) - \sum_{j: s_j \neq y_i} q(\mathbf{u_j^T x_i} - v_j) \right).
\end{align*}
Since $q$ is convex, $q(\mathbf{u_j^T x_i} - v_j)$ is also convex \citep{boyd2004convex}. As the loss function $L$ is convex, we finally conclude from Proposition \ref{prop:dc_properties} that the objective function is DC. Moreover, since the constraint $\mathbf{v} \geq \epsilon$ is convex, we conclude that (P) is a DC optimization problem. 

\subsection{Proof of Proposition \ref{prop:dc_decomposition}}

We now suppose that $L(x) = \max ( 0, 1 - x )$, and derive the DC form of the objective function. We have:
\noindent
\[
\begin{array}{lll}
& \multicolumn{2}{l}{\displaystyle \sum_{i=1}^m L \Bigl( y_i \displaystyle \sum_{j=1}^N s_j q(\mathbf{u_j^T x_i} - v_j) \Bigr)} \\
= & \multicolumn{2}{l}{\displaystyle \sum_{i=1}^m \max \Bigl(0, 1 + \displaystyle \sum_{j: s_j \neq y_i} q(\mathbf{u_j^T x_i} - v_j) - \displaystyle \sum_{j: s_j = y_i} q(\mathbf{u_j^T x_i} - v_j) \Bigr)} \\
= & \displaystyle \sum_{i=1}^m \max \Bigl( & \displaystyle \sum_{j: s_j = y_i} q(\mathbf{u_j^T x_i} - v_j) - \displaystyle \sum_{j: s_j = y_i} q(\mathbf{u_j^T x_i} - v_j), \\
& & 1 + \displaystyle \sum_{j: s_j \neq y_i} q(\mathbf{u_j^T x_i} - v_j) - \displaystyle \sum_{j: s_j = y_i} q(\mathbf{u_j^T x_i} - v_j) \Bigr) \\
= & \displaystyle \sum_{i=1}^m \max \Bigl( & \displaystyle \sum_{j: s_j = y_i} q(\mathbf{u_j^T x_i} - v_j), 1 + \displaystyle \sum_{j: s_j \neq y_i} q(\mathbf{u_j^T x_i} - v_j) \Bigr) \\
- & \multicolumn{2}{l}{\displaystyle \sum_{i=1}^m \displaystyle \sum_{j: s_j = y_i} q(\mathbf{u_j^T x_i} - v_j).}
\end{array}
\]

The objective function of (P) can therefore be written as $g-h$, with:
\[
\begin{array}{lll}
g & = \frac{\nu}{2} \| \mathbf{v} \|_2^2 + \displaystyle \sum_{i=1}^m \max \Bigl( & \displaystyle \sum_{j: s_j = y_i} q(\mathbf{u_j^T x_i} - v_j), \\
& & 1 + \displaystyle \sum_{j: s_j \neq y_i} q(\mathbf{u_j^T x_i} - v_j) \Bigr), \\
h & \multicolumn{2}{l}{= \displaystyle \sum_{i=1}^m \displaystyle \sum_{j: s_j = y_i} q(\mathbf{u_j^T x_i} - v_j),}
\end{array}
\]
where $g$ and $h$ are convex functions.

\section*{Acknowledgments}

The authors would like to thank the associate editor and the anonymous reviewers for their valuable comments and references that helped to improve the quality of this paper.

\bibliographystyle{spbasic}
\bibliography{reference.bib}

\end{document}